\documentclass{article}

\usepackage[preprint,nonatbib]{neurips_2020}

\usepackage[utf8]{inputenc} 
\usepackage[T1]{fontenc}    
\usepackage{hyperref}       
\usepackage{url}            
\usepackage{booktabs}       
\usepackage{amsfonts}       
\usepackage{amsmath}
\usepackage{nicefrac}       
\usepackage{microtype}      
\usepackage{graphicx}
\usepackage{subcaption}
\usepackage{multirow}
\usepackage{algorithm}
\usepackage{algorithmicx}
\usepackage{algpseudocode}

\usepackage{color}
\usepackage{soul}

\title{QuantNet: Transferring Learning \\ Across Trading Strategies}

\newif\ifshowcomments
\showcommentstrue

\author{%
Adriano Koshiyama\thanks{Corresponding author}, Stefano B. Blumberg, Nick Firoozye and Philip Treleaven \\
	University College London\\
	\texttt{\{adriano.koshiyama.15, stefano.blumberg.17, n.firoozye, p.treleaven\}@ucl.ac.uk} \\
  \And
	Sebastian Flennerhag \\
	University of Manchester \\
	\texttt{sebastian.flennerhag@postgrad.manchester.ac.uk} \\
}

\begin{document}

\maketitle


\begin{abstract}
	Systematic financial trading strategies account for over 80\% of trade volume in equities and a large chunk of the foreign exchange market. In spite of the availability of data from multiple markets, current approaches in trading rely mainly on learning trading strategies per individual market. In this paper, we take a step towards developing fully end-to-end global trading strategies that leverage systematic trends to produce superior market-specific trading strategies. We introduce \emph{QuantNet}: an architecture that learns market-agnostic trends and use these to learn superior market-specific trading strategies. Each market-specific model is composed of an encoder-decoder pair. The encoder transforms market-specific data into an abstract latent representation that is processed by a global model shared by all markets, while the decoder learns a market-specific trading strategy based on both local and global information from the market-specific encoder and the global model. QuantNet uses recent advances in transfer and meta-learning, where market-specific parameters are free to specialize on the problem at hand, whilst market-agnostic parameters are driven to capture signals from all markets. By integrating over idiosyncratic market data we can learn general transferable dynamics, avoiding the problem of overfitting to produce strategies with superior returns. We evaluate QuantNet on historical data across 3103 assets in 58 global equity markets. Against the top performing baseline, QuantNet yielded 51\% higher Sharpe and 69\% Calmar ratios. In addition we show the benefits of our approach over the non-transfer learning variant, with improvements of 15\% and 41\% in Sharpe and Calmar ratios. Code available in appendix.
\end{abstract}

\section{Introduction}\label{sec:introduction}

Systematic financial trading strategies account for over 80\% of trade volume in equities, a large chunk of the foreign exchange market, and are responsible to risk manage approximately \$500bn in assets under management \cite{avramovic2017we, Allenbridge}. High-frequency trading firms and e-trading desks in investment banks use many trading strategies, ranging from simple moving-averages and rule-based systems to more recent machine learning-based models \cite{heaton2017deep,de2018advances,molyboga2018portfolio,thomann2019factor,koshiyama2019avoiding,hiew2019bert,gu2020empirical}. 

Despite the availability of data from multiple markets, current approaches in trading strategies rely mainly on strategies that treat the relationships between different markets separately \cite{ghosn1997multi,bitvai2015day,heaton2017deep,de2018advances,zhang2019high,koshiyama2019avoiding,gu2020empirical}.  By considering each market in isolation, they fail to capture inter-market dependencies \cite{kenett2012correlations,raddant2016interconnectedness} like contagion effects and global macro-economic conditions that are crucial to accurately capturing market movements that allow us to develop robust market trading strategies. Furthermore, treating each market as an independent problem prevents effective use of machine learning since data scarcity will cause models to overfit before learning useful trading strategies \cite{Paper:RomanoStepwise:2005,Paper:HarveyBacktesting:2015,de2018advances,koshiyama2019avoiding}.

Commonly-used techniques in machine learning such as transfer learning \cite{caruana1997multitask,pan2009survey,zhang2019transfer,blumberg2019}, and multi-task learning \cite{caruana1997multitask,gibiansky2017deep,blumberg2019} could be used to handle information from multiple markets. However, combining these techniques is not immediately evident because these approaches often presume one task (market) as the main task and while others are auxiliary. When faced with several equally essential tasks, a key problem is how to assign weights to each market loss when constructing a multi-task objective. In our approach to end-to-end learning of global trading strategies, each market carries equal weight. This poses a challenge because (a) markets are non-homogeneous (e.g., size, trading days) and can cause interference during learning (e.g., errors from one market dominating others); (b) the learning problem grows in complexity with each market, necessitating larger models that often suffer from overfitting \cite{el2017scalable,he2019efficient} which is a notable problem in financial strategies \cite{Paper:RomanoStepwise:2005,Paper:HarveyBacktesting:2015,de2018advances,koshiyama2019avoiding}.

\begin{figure}[t!]
	\centering
	\includegraphics[width=\textwidth]{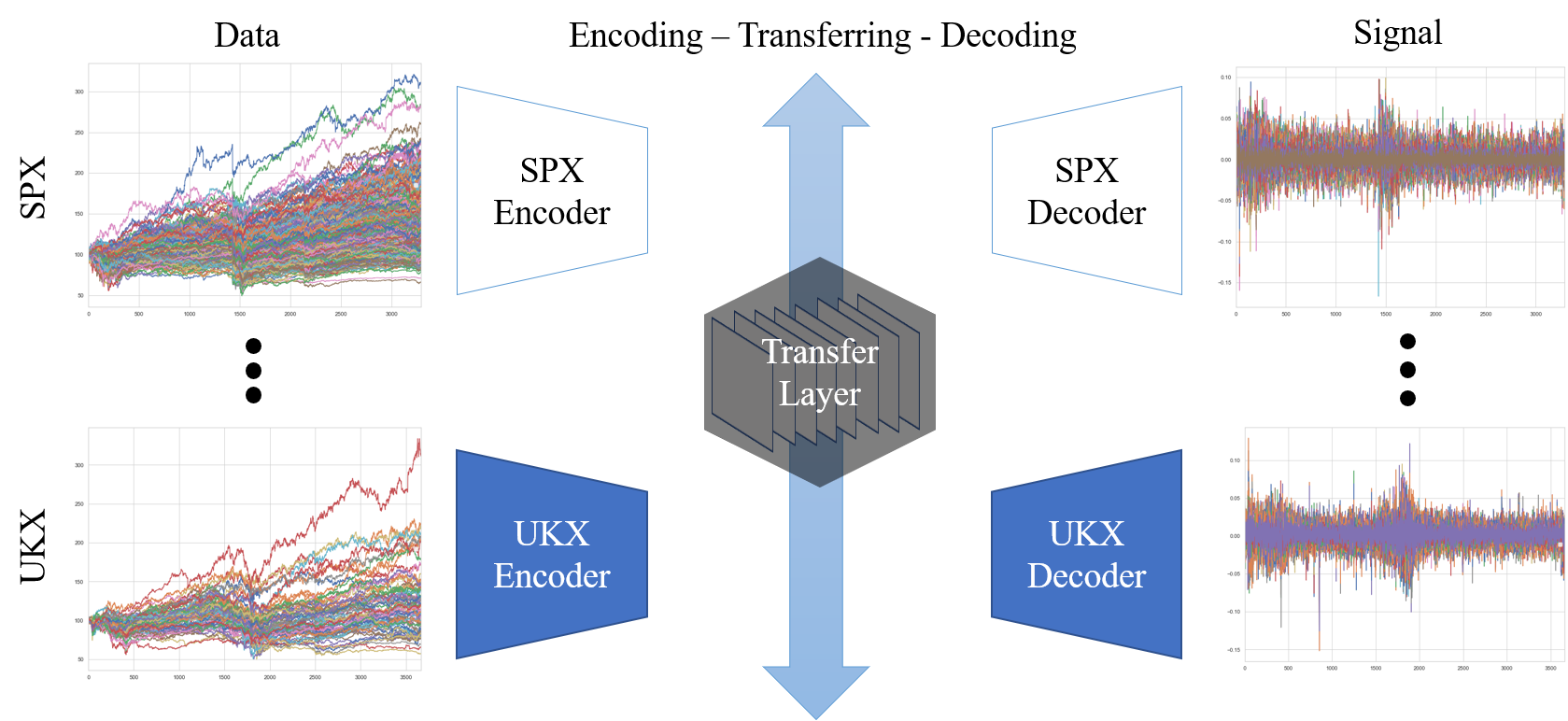}
	\caption{QuantNet workflow: from market data to decoding/signal generation.}
	\label{fig:QuantNet}
\end{figure}

In this paper, we take a step towards overcoming these challenges and develop a full end-to-end learning system for global financial trading. We introduce \emph{QuantNet}: an architecture that learns market-agnostic trends and uses them to learn superior market-specific trading strategies. Each market-specific model is composed of an encoder-decoder pair (Figure \ref{fig:QuantNet}). The encoder transforms market-specific data into an abstract latent representation that is processed by a global model shared by all markets, while the decoder learns a trading strategy based on the processed latent code returned by the global model. QuantNet leverages recent insights from transfer and meta-learning that suggest market-specific model components benefit from having separate parameters while being constrained by conditioning on an abstract global representation \cite{Paper:RuderTransfer:2019}. Furthermore, by incorporating multiple losses into a single network, our approach increases network regularization and avoids overfitting, as in \cite{lee2015,wang2015,blumberg2019}. We evaluate QuantNet on historical data across 3103 assets in 58 global equity markets. Against the best performing baseline Cross-sectional momentum \cite{jegadeesh1993returns, baz2015dissecting}, QuantNet yields 51\% higher Sharpe and 69\% Calmar ratios. Also, we show the benefits of our approach, which yields improvements of 15\% and 41\% in Sharpe and Calmar ratios, respectively, over the comparable non-transfer learning variant.  

Our key contributions are: (\textbf{i}) a novel architecture for transfer learning across financial trading strategies; (\textbf{ii}) a novel learning objective to facilitate end-to-end training of trading strategies; and (\textbf{iii}) demonstrate that QuantNet can achieve significant improvements across global markets with an end-to-end learning system. To the best of our knowledge, this is the first paper that studies transfer learning as a means of improving end-to-end large scale learning of trading strategies. 

\section{Related Work}

\textbf{Trading Strategies with Machine Learning}  Machine learning-based trading strategies have previously been explored in the setting of supervised learning \cite{heaton2017deep,aggarwal2017deep,fischer2018deep,sezer2020financial,gu2020empirical} and reinforcement learning \cite{pendharkar2018trading,li2019application,gao2020application,zhang2020deep}, nowadays with a major emphasis on deep learning methods. Broadly, these works differ from our proposed method as they do not use inter-market knowledge transfer and (in the case of methods based on supervised learning) tend to forecast returns/prices rather than to generate end-to-end trading signals. We provide an in-depth treatment in Section \ref{sec:method:prelim1}.

\textbf{Transfer Learning} Transfer learning is a well-established method in machine learning \cite{caruana1997multitask,pan2009survey,zhang2019transfer}. It has been used in computer vision, medicine, and natural language processing \cite{blumberg2018deeper,kornblith2019better,radford2019language}. In financial systems, this paradigm has primarily been studied in the context of applying unstructured data, such as social media, to financial predictions \cite{araci2019finbert,hiew2019bert}. 
In a few occasions such methodologies have been applied to trading, usually combined with reinforcement learning and to a very limited pool of assets \cite{jeong2019improving}. We provide a thorough review of the area in appendix \ref{literature}.

The simplest and most common form of transfer learning pre-trains a model on a large dataset, hoping that the pre-trained model can be fine-tuned to a new task or domain at a later stage \cite{devlin2018bert,liu2019roberta}. While simple, this form of knowledge transfer assumes new tasks are similar to previous tasks, which can easily fail in financial trading where markets differ substantially. Our method instead relies on multi-task transfer learning \cite{Caruana1993multitasklearning,baxter1995learning,caruana1997multitask,ruder2017overview}. While this approach has previously been explored in a financial context \cite{ghosn1997multi,bitvai2015day}, prior works either use full parameter sharing or share all but the final layer of relatively simple models. In contrast, we introduce a novel architecture that relies on encoding market-specific data into representations that pass through a global bottleneck for knowledge transfer. We provide a detailed discussion in Section \ref{sec:method:prelim2}.

\section{QuantNet}
\vspace{-0.15cm}
We begin by reviewing end-to-end learning of financial trading in section \ref{sec:method:prelim1} and relevant forms of transfer learning in section \ref{sec:method:prelim2}. We present our proposed architecture in section \ref{sec:method:quantnet}. 

\subsection{Preliminaries: Learning Trading Strategies}\label{sec:method:prelim1}

A financial market $M = (a^1, \ldots, a^n)$ consists of a set of $n$ assets $a^j$; at each discrete time step $t$ we have access to a vector of excess returns $\boldsymbol{r}_t = (r^1_t, \ldots, r^n_t) \in \mathbb{R}^{n}$. The goal of a trading strategy $f$, parametrized by $\theta$, is to map elements of a history $R_{m:t} = (\boldsymbol{r}_{t-m}, \ldots, \boldsymbol{r}_t)$ into a set of \emph{trading signals} $\boldsymbol{s}_t = (s^1_t, \ldots, s^n_t) \in \mathbb{R}^{n}$; $\boldsymbol{s}_t = f_{\theta}(R_{m:t})$. These signals constitute a market portfolio: a trader would buy one unit of an asset if $s^j_{t} = 1$, sell one unit if $s^j_t = -1$, and close a position if $s^j_t = 0$; any value in between $(-1, 1)$ implies that the trader is holding/shorting a fraction of an asset. The goal is to produce a sequence of signals that maximize risk-adjusted returns. The most common approach is to model $f$ as a moving average parametrized by weights $\theta = (W_1, \ldots, W_m)$, $W_i \in \mathbb{R}
^{n \times n}$, where weights typically decreases exponentially or are hand-engineered and remain static \cite{BarclayHedge,avramovic2017we,firoozye2019optimal}; 
\begin{equation}
\boldsymbol{s}_{t}^{\text{ma}} = \tau(f^{\text{ma}}_\theta(R_{t-m:t})) = \tau\left(\sum_{k=t-m}^{t} W_k \boldsymbol{r}_k\right), \quad \tau: \mathbb{R}^n \rightarrow [-1, 1]^n. 
\end{equation}
More advanced models rely on recurrence to form an abstract representation of the history up to time $t$; either by using Kalman filtering, Hidden Markov Models (HMM), or Recurrent Neural Networks (RNNs) \cite{zhang2019high,fischer2018deep}. For our purposes, having an abstract representation of the history will be crucial to facilitate effective knowledge transfer, as it captures each market's dynamics, thereby allowing QuantNet to disentangle general and idiosyncratic patterns. We use the Long Short-Term Memory (LSTM) network \cite{hochreiter1997long,gers1999learning}, which is a special form of an RNN. LSTMs have been recently explored to form the core of trading strategy systems \cite{fischer2018deep,hiew2019bert,sezer2020financial}. For simplicity, we present the RNN here and refer the interested reader to \cite{hochreiter1997long,gers1999learning} or appendix \ref{architecture}. The RNN is defined by introducing a recurrent operation that updates a hidden representation $\boldsymbol{h}$ recurrently conditional on the input $\boldsymbol{r}$: 
\begin{equation}\label{eq:rnn}
\boldsymbol{s}^{\text{RNN}}_{t} = \tau(W_s \boldsymbol{h}_t + \boldsymbol{b}_s), \quad \boldsymbol{h}_t = f^{\text{RNN}}_{\theta}(\boldsymbol{r}_{t-1}, \boldsymbol{h}_{t-1}) = \sigma \left( W_r \boldsymbol{r}_{t-1} + W_h \boldsymbol{h}_{t-1} + \boldsymbol{b}\right),
\end{equation}
where $\sigma$ is an element-wise activation function and $\theta = (W_r, W_h, \boldsymbol{b})$ parameterize the RNN. The LSTM is similarly defined but adds a set of gating mechanisms to enhance the memory capacity and gradient flow through the model. To learn the parameters of the RNN, we use truncated backpropagation through time (TBPTT; \cite{TBPTT}), which backpropagates a loss $\mathcal{L}$ through time to the parameters of the RNN, truncating after $K$ time steps. 

\subsection{Preliminaries: Transfer Learning}\label{sec:method:prelim2}

Transfer learning \cite{caruana1997multitask,pan2009survey,ruder2017overview,zhang2019transfer} embodies a set of techniques for sharing information obtained on one task, or market, when learning another task (market). In the simplest case of pre-training \cite{devlin2018bert,liu2019roberta}, for example, we would train a model in a market $M_2$ by initializing its parameters to the final parameters obtained in market $M_1$. While such pre-training can be useful, there is no guarantee that the parameters obtain on task $M_1$ will be useful for learning task $M_2$. 

In multi-task transfer-learning, we have a set of $\mathcal{M} = (M_1, \ldots, M_N)$ markets that we aim to learn simultaneously. The multi-task literature often presumes one $M_i$ is the main task, and all others are auxiliary---their sole purpose is to improve final performance on $M_i$ \cite{caruana1997multitask,gibiansky2017deep}. A common problem in multi-task transfer is therefore how to assign weights $w_i$ to each market-specific loss $\mathcal{L}^i$ when setting the multi-task objective $\mathcal{L} = \sum_i w_ i \mathcal{L}^i$. This is often a hyper-parameter that needs to be tuned \cite{ruder2017overview}.


As mentioned above, this poses a challenge in our case as markets are typically not homogeneous. 
Instead, we turn to sequential multi-task transfer learning \cite{Paper:RuderTransfer:2019}, which learns one model $f^i$ per market $M_i$, but partition the parameters of the model into a set of market-specific parameters $\theta^i$ and a set of market-agnostic parameters $\phi$. In doing so, market-specific parameters are free to specialize on the problem at hand, while market-agnostic parameters capture signals from all markets. However, in contrast to standard approaches to multi-task transfer learning, which either share all parameters, a set from the final layer(s) or share no parameters, we take inspiration from recent advances in meta-learning \cite{lee2018gradient,zintgraf2018fast,flennerhag2019meta}, which shows that more flexible parameter-sharing schemes can reap a greater reward. In particular, \emph{interleaving} shared and market-specific parameters can be seen as learning both shared representation and a shared optimizer \cite{flennerhag2019meta}. 

We depart from previous work by introducing an encoder-decoder setup \cite{Cho:2014ed} within financial markets. Encoders learn to represent market-specific information, such as internal fiscal and monetary conditions, development stage, and so on, while a global shared model learns to represent market-agnostic dynamics such as global economic outlook, contagion effects (via financial crises). The decoder uses these sources of information to produce a market-specific trading strategy. 
With these preliminaries, we now turn to QuantNet, our proposed method for end-to-end multi-market financial trading.


\subsection{QuantNet}\label{sec:method:quantnet}

\begin{figure}[t!]
	\centering
	\includegraphics[width=\textwidth]{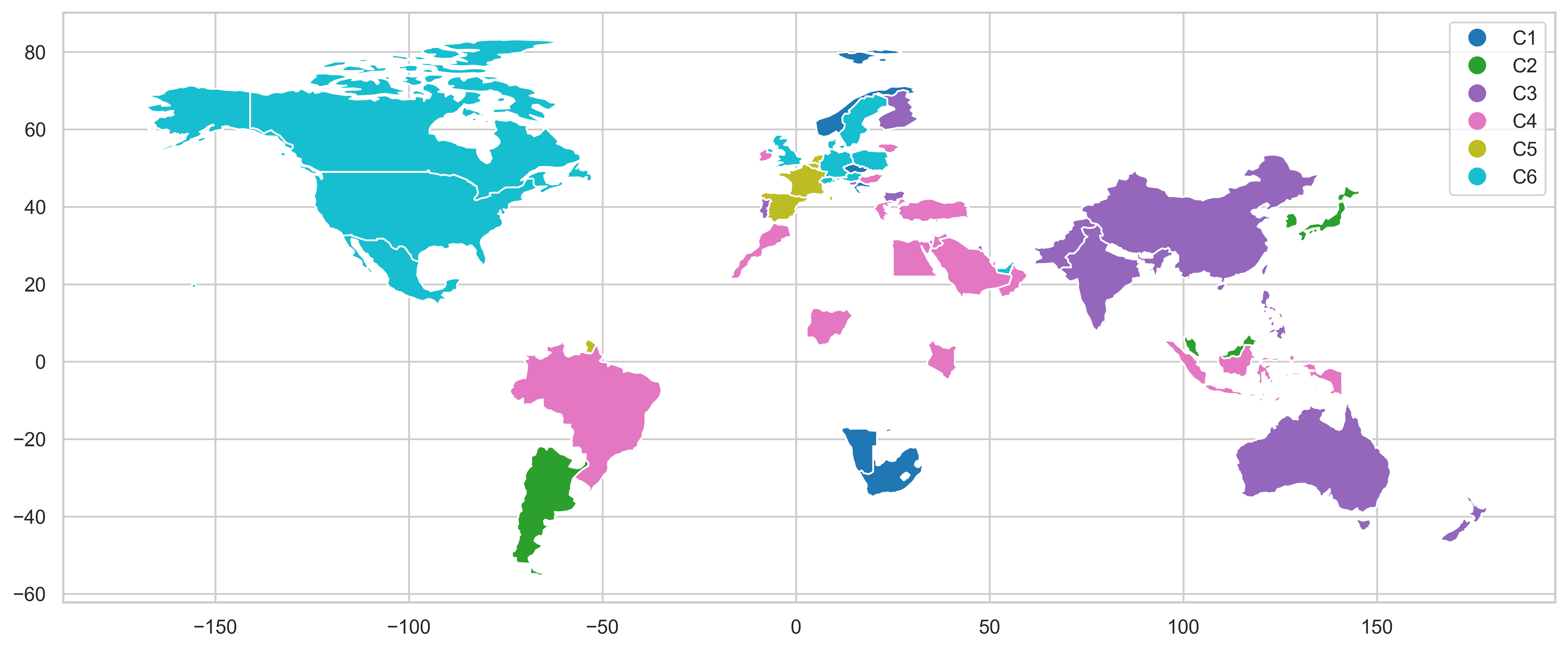}
	\caption{World map depicting the different clusters formed from the scores of QuantNet encoder. For visualization purposes, we have picked the market with the biggest market capitalization to represent the country in the cluster.}
	\label{fig:worldmap}
\end{figure}

\paragraph{Architecture} Figure \ref{fig:QuantNet} portrays the QuantNet architecture. In QuantNet, we associate each market $M_i$ with an encoder-decoder pair, where the encoder $\operatorname{enc}^i$ and the decoder $\operatorname{dec}^i$ are both LSTMs networks. Both models maintain a separate hidden state, $\boldsymbol{e}^i$ and $\boldsymbol{d}^i$, respectively. When given a market return vector $\boldsymbol{r}^i$, the encoder produces an encoding $\boldsymbol{e}^i$ that is passed onto a market-agnostic model $\omega$, which modifies the market encoding into a representation $\boldsymbol{z}^i$:
\begin{equation}\label{eq:encode}
\boldsymbol{z}^i_t = \omega(\boldsymbol{e}^i_t), \quad \text{where} \quad
\boldsymbol{e}^i_t = \operatorname{enc}^i(\boldsymbol{r}^i_{t-1}, \boldsymbol{e}^i_{t-1}).
\end{equation}
Because $\omega$ is shared across markets, $\boldsymbol{z}^i$ reflects market information from market $M^i$ while taking global information (as represented by $\omega$) into account. This bottleneck enforces local representations that are aware of global dynamics, and so we would expect similar markets to exhibits similar representations \cite{mikolov2013distributed}. We demonstrate this empirically in Figure \ref{fig:worldmap}, which shows how each market is being represented internally by QuantNet. We apply hierarchical clustering on hidden representation from the encoder (see also dendrogram in appendix \ref{dendrogram}) using six centroids. 
We observe clear geo-economical structure emerging from QuantNet -- without it receiving \emph{any} such geographical information. $C5$ consist mainly of small European equity markets (Spain, Netherlands, Belgium, and France) -- all neighbors; $C6$ encompass developed markets in Europe and Americas, such as United Kingdom, Germany, US, and their respective neighbors Austria, Poland, Switzerland, Sweden, Denmark, Canada, and Mexico. Other clusters are more refined: $C2$ for instance contains most developed markets in Asia like Japan, Hong Kong, Korea, and Singapore, while $C3$ represents Asia and Pacific emerging markets: China, India, Australia, and some respective neighbors (New Zealand, Pakistan, Philippines, Taiwan). 

We experiment with different functional forms for $\omega$; adding complexity to $\omega$ can allow more sophisticated representations, but simpler architectures enforce an information bottleneck \cite{yu2019understanding}. Indeed we find experimentally that a simple linear layer works better than an LSTM (see Appendix \ref{ablation}), which is in line with general wisdom on encoder-decoder architectures \cite{cho2014learning,Cho:2014ed}. 

Given a representation $\boldsymbol{z}^i_t$, we produce a market-specific trading strategy by decoding this abstract representation into a hidden market-specific state $\boldsymbol{d}^i_t$; this state represents the trading history in market $M^i$, along with the history of global dynamics, and is used learn a market-specific trading strategy
\begin{equation}\label{eq:quantnet}
\boldsymbol{s}^{i}_t = f^i(\boldsymbol{d}^i_t) = \tanh(W^i \boldsymbol{d}^i_t + \boldsymbol{b}^i),
\quad \text{where} \quad
\boldsymbol{d}^i_t = \operatorname{dec}^i(\boldsymbol{z}^i_t, \boldsymbol{d}^i_{t-1}).
\end{equation}
While $f^i$ can be any model, we found a simple linear layer sufficient due to the expressive capacity of the encoder-decoder pair. For a well-behaved, non-leveraged trading strategy we chose $\tanh$ as our activation function, which bounds the trading signal $\boldsymbol{s}^i_t \in (-1, 1)^n$ \cite{Book:AcarTradRules:2002,voit2013statistical}. 

From Eq. \ref{eq:quantnet}, we can see how transfer learning affects both trading and learning. During trading, by processing a market encoding $\boldsymbol{e}^i_t$ through a shared global layer $\omega$, we impose a bottleneck such that any trading strategy is forced to act on the globally conditioned information in $\boldsymbol{z}^i_t$. During learning, market-specific trading is unrestricted in its parameter updates, but gradients are implicitly modulated through the conditioned input. This is particularly true for the encoder, which must service its corresponding decoder by passing through the global layer $\omega$. Concretely, given a market loss function $\mathcal{L}^i$ with error signal $\delta^i = d \mathcal{L} / d \boldsymbol{s}^i$, market-specific gradients are given by
\begin{equation}\label{eq:grads1}
\nabla_{\theta^i_{\text{dec}^i}} \mathcal{L}^i(\boldsymbol{s}^i_t)
=
\delta^i_t
\frac{\partial \boldsymbol{s}^i_t}{\partial \boldsymbol{d}^i_t} 
\frac{\partial \operatorname{dec}^i}{\partial \theta^i_{\text{dec}^i}} (\boldsymbol{z}^i_t; \boldsymbol{d}^i_{t-1}),
\quad
\nabla_{\theta^i_{\text{enc}^i}} \mathcal{L}^i(\boldsymbol{s}^i_t)
=
\delta^i_t
\frac{\partial \boldsymbol{s}^i_t}{\partial \boldsymbol{d}^i_t} 
\frac{\partial \boldsymbol{d}^i_t}{\partial \boldsymbol{z}^i_t} 
\frac{\partial \boldsymbol{z}^i_t}{\partial \boldsymbol{e}^i_t} 
\frac{\partial \operatorname{enc}^i}{\partial \theta^i_{\text{enc}^i}} (\boldsymbol{r}^i_t; \boldsymbol{e}^i_{t-1}).
\end{equation}
The gradient of the decoder is largely free but must adapt to the representation produced by the encoder and the global model. These, in turn, are therefore influenced by what representations are useful for the decoder. In particular, the gradient of the encoder must pass through the global model, which acts as a preconditioner of the encoder parameter gradient, thereby encoding an optimizer \cite{Flennerhag2020warpgrad}. Finally, to see how global information gets encoded in the global model, under a multi-task loss, its gradients effectively integrate out idiosyncratic market correlations:
\begin{equation}\label{eq:grads2}
\nabla_{\phi} \mathcal{L}(\boldsymbol{s}^1_t, \ldots, \boldsymbol{s}^n_t)
=
\sum_{i=1}^n
\delta^i_t
\frac{\partial \boldsymbol{s}^i_t}{\partial \boldsymbol{d}^i_t} 
\frac{\partial \boldsymbol{d}^i_t}{\partial \boldsymbol{z}^i_t} 
\frac{\partial \omega}{\partial \phi} (\boldsymbol{e}^i_t).
\end{equation}
\paragraph{Learning Objective} To effectively learn trading strategies with QuantNet, we develop a novel learning objective based on the Sharpe ratio  \cite{sharpe1994sharpe,Paper:BaileySharpeFrontier:2012,Paper:HarveyBacktesting:2015}. Prior work on financial forecast tends to rely on Mean Squared Error (MSE) \cite{heaton2017deep,hiew2019bert,gu2020empirical}, as does most work on learning trading strategies. A few works have instead considered other measurements \cite{fischer2018deep,zhang2019high,zhang2020deep}. In particular, there are strong theoretical and empirical reasons for considering the Sharpe ratio instead of MSE -- in fact, MSE minimization is a necessary, but not sufficient condition to maximize the profitability of a trading strategy \cite{bengio1997using,Book:AcarTradRules:2002,Paper:KoshiyamaCovPen:2019}. Since some assets are more volatile than others, the Sharpe ratio helps to discount the optimistic average returns by taking into account the risk faced when traded those assets. Also, it is widely adopted by quantitative investment strategists to rank different strategies and funds \cite{sharpe1994sharpe,Paper:BaileySharpeFrontier:2012,Paper:HarveyBacktesting:2015}. 

To compute each market Sharpe ratio at a time $t$, truncated to backpropagation through time for $k$ steps (to $t-k$), considering excess daily returns, we first compute the per-asset Sharpe ratio 
\begin{equation}\label{eq:sharpe}
\rho^i_{t,j} = \left.\left(\mu^i_{t-k:t, j} \right) \right/ \left(\sigma^i_{t-k:t, j} \right) \cdot \sqrt{252},
\end{equation}
where $\mu^i_{t, j}$ is the average return of the strategy for asset $j$ and $\sigma^j_{t}$ is its respective the standard deviation. The $\sqrt{252}$ factor is included in computing the annualized Sharpe ratio. The market loss function and the QuantNet objective are given by averaging over assets and markets, respectively:
\begin{equation}\label{eq:loss}
\mathcal{L}\left({\{\boldsymbol{s}^i_{t-k:t}, \boldsymbol{r}^i_{t-k:t}\}}_{i=1}^N\right) = \frac1N \sum_{i=1}^N \mathcal{L}^i(\boldsymbol{s}^i_{t-k:t}, \boldsymbol{r}^i_{t-k:t}),
\quad
\mathcal{L}^i(\boldsymbol{s}^i_{t-k:t}, \boldsymbol{r}^i_{t-k:t}) = \frac1n \sum_{j=1}^n \rho^i_{t,j},
\end{equation}
\paragraph{Training} To train QuantNet, we use stochastic gradient descent. To obtain a gradient update, we first sample mini-batches of $m$ markets from the full set $\mathcal{M} = (M_1, \ldots, M_N)$ to obtain an empirical expectation over markets. Given these, we randomly sample a time step $t$ and run the model from $t-k$ to $t$, from which we obtain market Sharpe ratios. Then, we compute QuantNet loss function and differentiate through time into all parameters. We pass this gradient to an optimizer, such as Adam, to take one step on the model's parameters. This process is repeated until the model converges.
\begin{algorithm}[t!]
{
	\caption{QuantNet Training}\label{training}
	\begin{algorithmic}[1]
		\Require{Markets $\mathcal{M} = (M_1, \ldots M_N)$}
		\Require{Backpropagation horizon $k$}
		\While{True}
		\State{Sample mini-batch $M$ of $m$ markets from $\mathcal{M}$}
		\State{Randomly select $t \in 1, \ldots, T$} 
		\State{Compute encodings $\boldsymbol{e}^i_{t-k:t}$, $\boldsymbol{z}^i_{t-k:t}$, and $\boldsymbol{d}^i_{t-k:t}$ for all $M_i \in M$ \hfill Eqs. \ref{eq:encode} and \ref{eq:quantnet}}
		\State{Compute signals $\boldsymbol{s}^i_{t-k:t}$ for all $M_i \in M$
			\hfill Eq. \ref{eq:quantnet}}
		\State{Compute Sharpe ratios $\rho^i_{t:j}$ for all assets $a^i_j \in M_i$ and markets $M_i \in M$
			\hfill Eq. \ref{eq:sharpe}}
		\State{Compute QuantNet loss $\mathcal{L}\left({\{\boldsymbol{s}^i_{t-k:t}, \boldsymbol{r}^i_{t-k:t}\}}_{i=1}^N\right) $
			\hfill Eq. \ref{eq:loss}}	
		\State{Update model parameters by truncated backpropagation through time
			\hfill
			Eqs. \ref{eq:grads1} and \ref{eq:grads2}}	
		\EndWhile
	\end{algorithmic}
}
\end{algorithm}
\vspace{-0.15cm}
\section{Results}
\vspace{-0.15cm}
This section assesses QuantNet performance compared to baselines and a \emph{No Transfer} strategy defined by a single LSTM of the same dimensionality as the decoder architecture (number of assets), as defined in Eq. \ref{eq:rnn}. Next section presents the main experimental setting, with the subsequent ones providing: (i) a complete comparison of QuantNet with other trading strategies; (ii) an in-depth comparison of QuantNet versus the best No Transfer strategy; and (iii) analysis on market conditions that facilitate transfer under QuantNet. We provide an ablation study and sensitivity analysis of QuantNet in appendix \ref{ablation}.

\subsection{Experimental Setting}

\paragraph{Datasets} Appendix \ref{datasets} provides a full table listing all 58 markets used. We tried to find a compromise between the number of assets and sample size, hence for most markets, we were unable to use the full list of constituents. We aimed to collect daily price data ranging from 03/01/2000 to 15/03/2019, but for most markets it starts roughly around 2010. Finally, due to restrictions from our Bloomberg license, we were unable to access data for some important equity markets, such as Italy and Russia.

\paragraph{Evaluation} Appendix \ref{benchmarking} provides full experimental details, including detailed descriptions of baselines and hyperparameter search protocols. 
We report results for trained models under best hyperparameters on validation sets; for each dataset we construct a training and validation set, where the latter consists of the last 752 observations of each time series (around 3 years). We have used 3 Month London Interbank Offered Rate in US Dollar as the reference rate to compute excess returns. We have also reported Calmar ratios, Annualized Returns and Volatility, Downside risk, Sortino ratios, Skewness and Maximum drawdowns \cite{young1991calmar,sharpe1994sharpe,eling2007does,rollinger2013sortino}. 

\subsection{Empirical Evaluation}

\paragraph{Baseline Comparison} Table \ref{Table-Global-Stats} present median and mean absolute deviation (in brackets) performance of the different trading strategies on 3103 stocks across all markets analysed. The best baseline is Cross-sectional Momentum (CS Mom), yielding a SR of 0.23 and CR of 0.14. QuantNet outperforms CS Mom, yielding 51\% higher SR and 69\% higher CR. No Transfer LSTM and Linear outperforms this baseline as well, but not to the same extent as QuantNet.

\begin{table}[h!]
	\caption{Median and mean absolute deviation (in brackets) performance on 3103 stocks across all markets analysed. TS Mom - Time series momentum and CS Mom - Cross-section momentum. We highlighted in bold only the metrics where a comparison can be made, like Sharpe ratios, Calmar ratios, Kurtosis, Skewness, and Sortino ratios.}
	\centering
	\scriptsize
	\label{Table-Global-Stats}
	\begin{tabular}{c|ccccccc}
		\hline
		\hline
Metric   & Buy   and hold & Risk   parity & TS Mom            & CS Mom            & No   Transfer LSTM & No   Transfer Linear & QuantNet         \\
\hline
Ann Ret  & 0.000020       & 0.000193      & 0.000050          & 0.000042          & 0.002508           & 0.001537             & 0.005377         \\
         & (0.13433)    & (0.00270)     & (0.00019)         & (0.00019)         & (0.07645)         & (0.08634)           & (0.02898)     \\
    \hline
Ann Vol  & 0.287515       & 0.001536      & 0.000290          & 0.000270          & 0.008552           & 0.007768             & 0.023665    \\
         & (0.10145)    & (0.00537)     & (0.00036)         & (0.00036)         & (0.13455)         & (0.14108)           & (0.04540)    \\
    \hline
CR       & 0.000040       & 0.095516      & 0.139599          & 0.143195          & 0.158987           & 0.169345             & \textbf{0.241255}   \\
         & (0.33583)     & (0.29444)     & (1.05288)         & (1.18751)         & (0.55762)         & (0.57170)           & (0.59968)     \\
    \hline
DownRisk & 0.202361       & 0.001076      & 0.000195          & 0.000178          & 0.005656           & 0.005124             & 0.015734    \\
         & (0.07042)    & (0.00361)      & (0.00024)         & (0.00025)         & (0.09223)         & (0.09553)            & (0.03291)     \\
    \hline
Kurt     & \textbf{5.918386}       & 6.165916      & 13.333863         & 18.112853         & 16.87256           & 15.73864             & 16.19961    \\
         & (10.2515)    & (13.9426)     & (19.2278)         & (24.4672)           & (30.2204)        & (31.0395)           & (24.7336)      \\
    \hline
MDD      & -0.419984      & -0.002935     & -0.000488         & -0.000444         & -0.014564          & -0.01286             & -0.03847       \\
         & (0.14876)    & (0.00987)     & (0.00082)         & (0.00081)         & (0.16724)         & (0.17820)             & (0.07881)   \\
    \hline
SR       & 0.000051       & 0.155560      & 0.226471          & 0.234583          & 0.304244           & 0.306572             & \textbf{0.354776}        \\
         & (0.42324)    & (0.42028)      & (0.40627)         & (0.41547)         & (0.51552)         & (0.51182)            & (0.57218)    \\
    \hline
Skew     & -0.087282      & -0.092218     & 0.427237          & \textbf{0.568364}          & 0.256736           & 0.171629             & 0.297182       \\
         & (0.82186)    & (0.96504)     & (1.28365)         & (1.60612)         & (1.77245)         & (1.74804)           & (1.66854)        \\
    \hline
SortR    & 0.217621       & 0.220335      & 0.333685          & 0.349124          & 0.443422        & 0.454035             & \textbf{0.52196}  \\
         & (0.59710)     & (0.61883)     & (1.02616) & (0.633953) & (0.78525)  & (0.86715)           & (1.02465) \\
       \hline
       \hline
	\end{tabular}
\end{table}

\begin{figure}[h!]
	\centering
	\includegraphics[width=\textwidth]{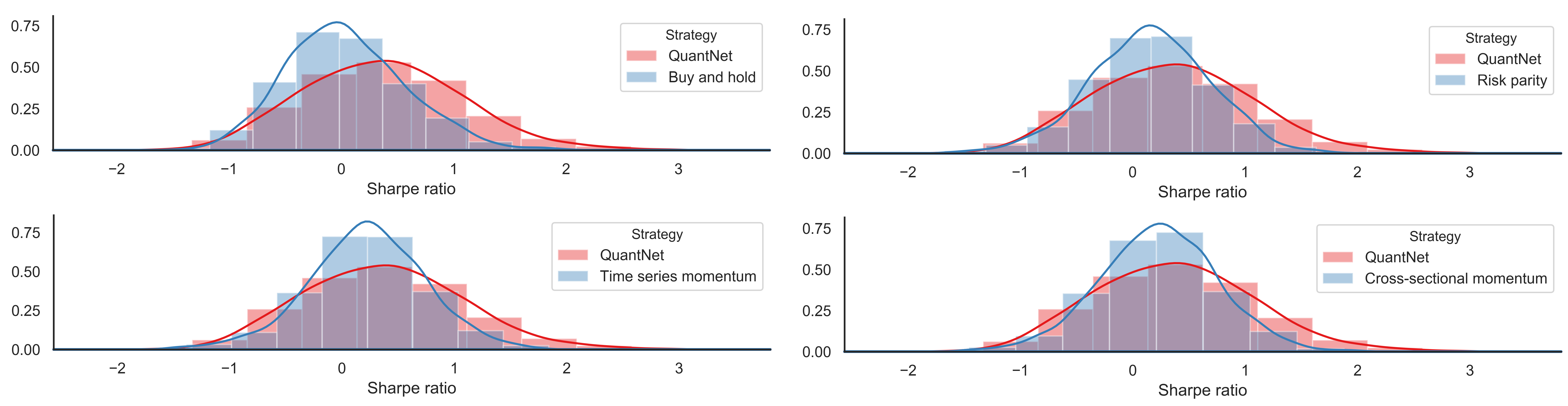}
	\caption{Histogram of Sharpe ratio contrasting QuantNet with baseline strategies.}
	\label{fig:hist-baselines}
\end{figure}

\paragraph{QuantNet vs No Transfer Linear}When comparing QuantNet and No Transfer Linear strategies performance (Table \ref{Table-Global-Stats}), we observe an improvement of about 15\% on SR and 41\% on CR. This improvement increases the number of assets yielding SRs above 1.0 from 432 to 583, smaller Downside Risk (DownRisk), higher Skew and Sortino ratios (SortR). Statistically, QuantNet significantly outperform No Transfer both in Sharpe ($W=2215630$, p-value $<$ 0.01) and Calmar ($W=2141782$, p-value $<$ 0.01) ratios. This discrepancy manifests in statistical terms, with the Kolmogorov-Smirnov statistic indicating that these distributions are meaningfully different ($KS=0.053$, p-value $<$ 0.01). 

\begin{figure}[h!]
	\centering
	\includegraphics[width=\columnwidth]{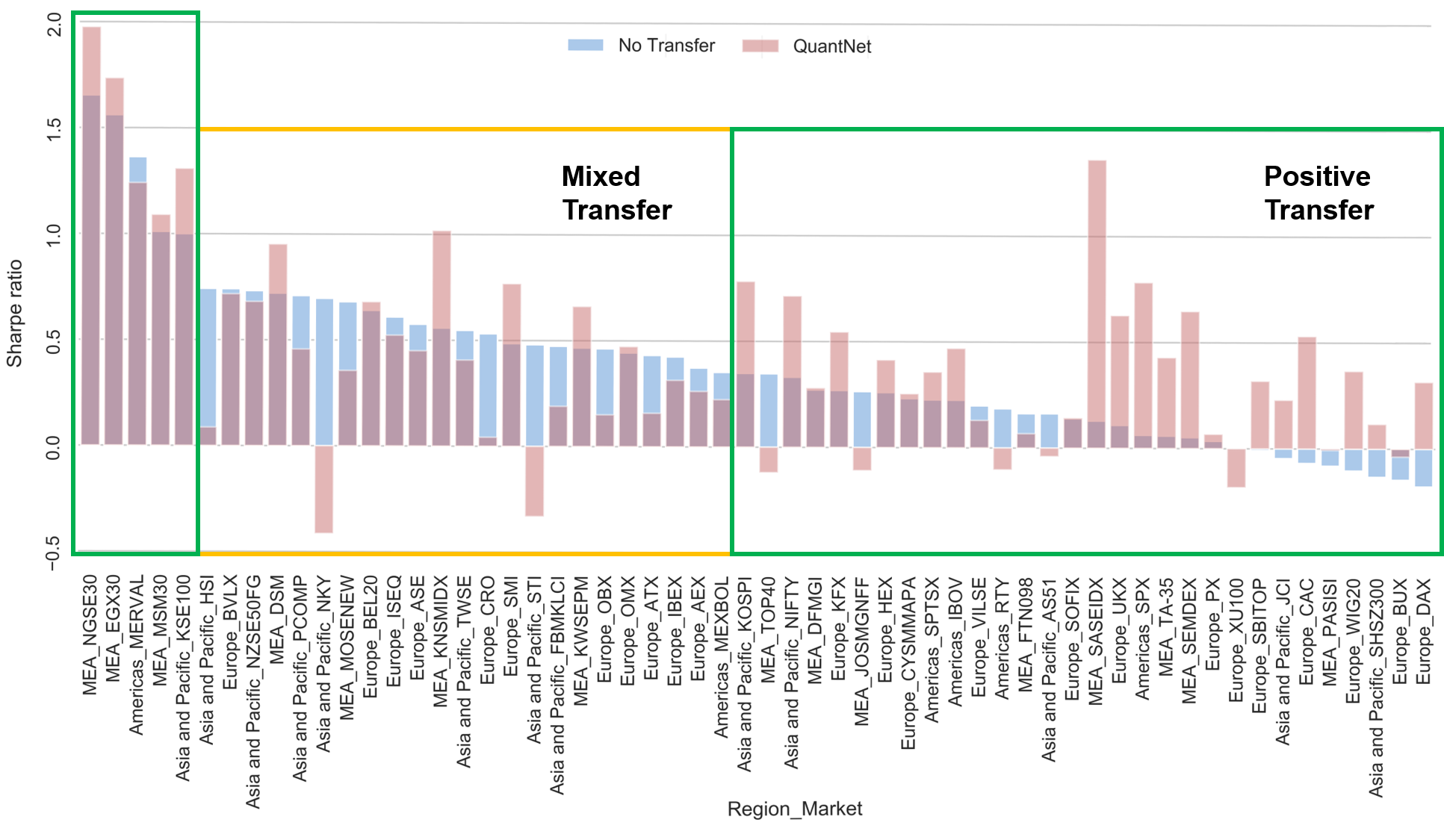}
	\caption{Average Sharpe ratios of QuantNet and No Transfer across 58 equity markets.} 
	\label{fig:barchart}
\end{figure}

\begin{figure}[h!]
	\centering
	\begin{subfigure}{.69\textwidth}
		\includegraphics[width=\textwidth]{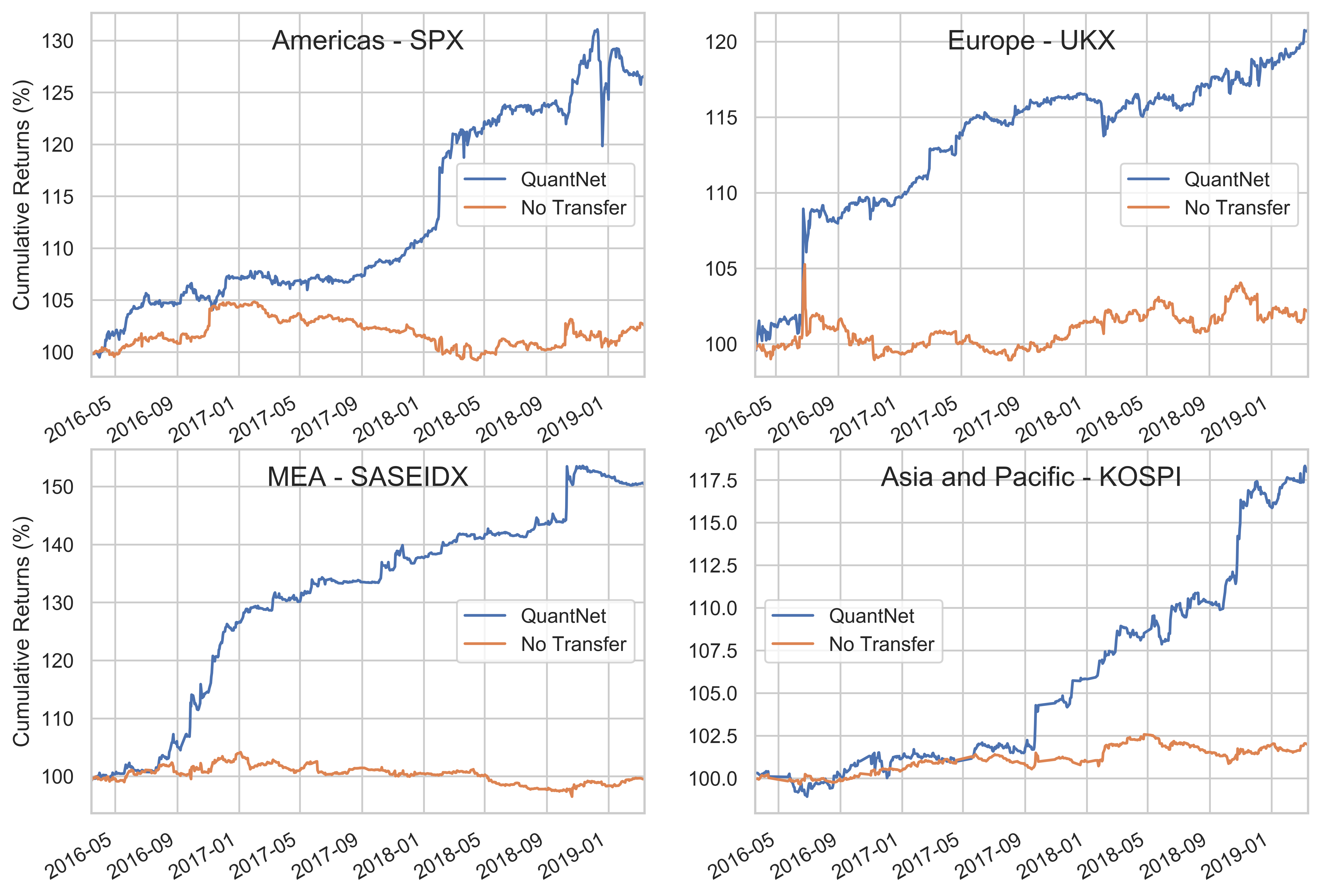}
		\caption{}
	\end{subfigure}
	\begin{subfigure}{.295\textwidth}
		\centering
		\includegraphics[width=\linewidth]{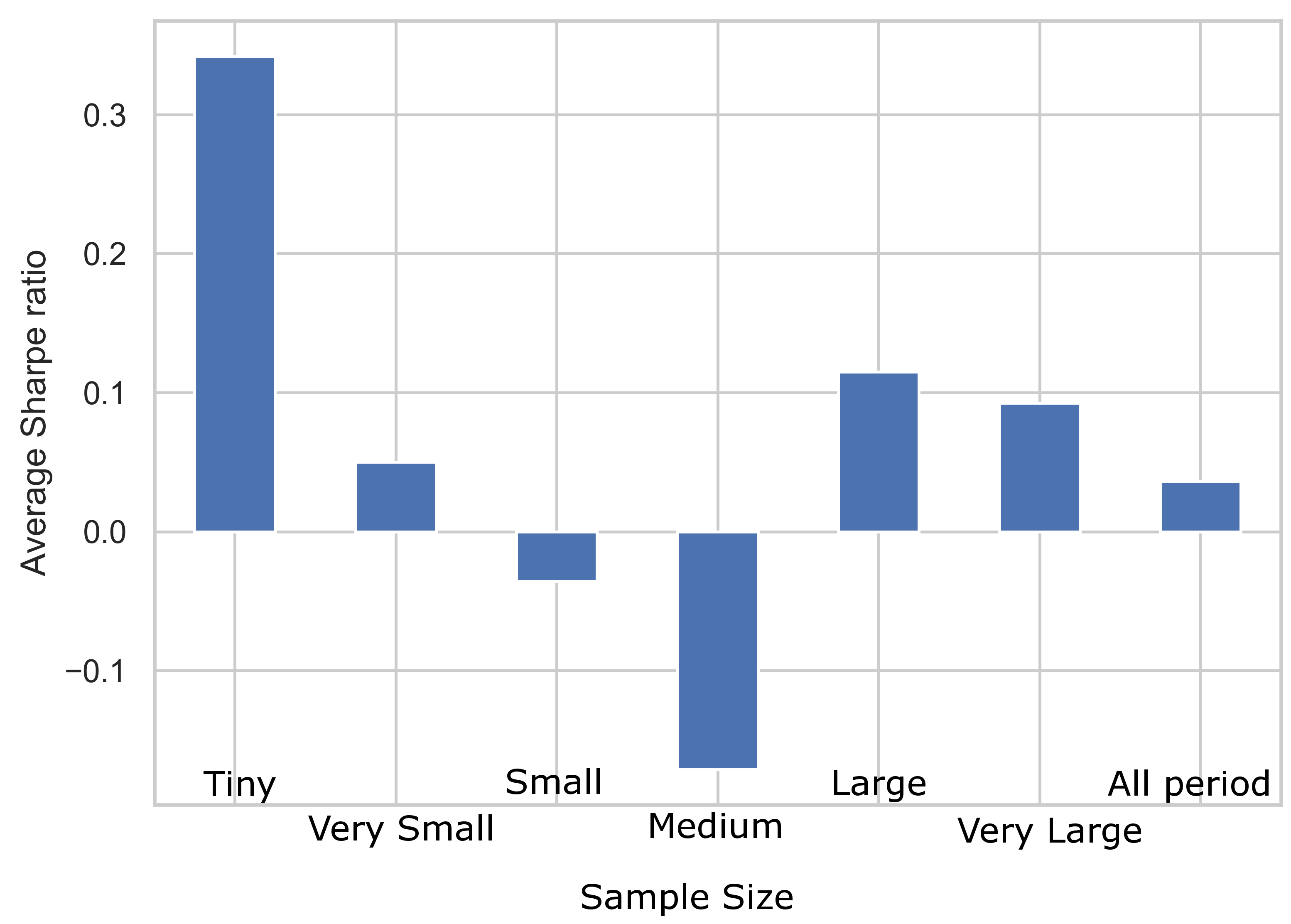}
		\caption{}
		\includegraphics[width=\linewidth]{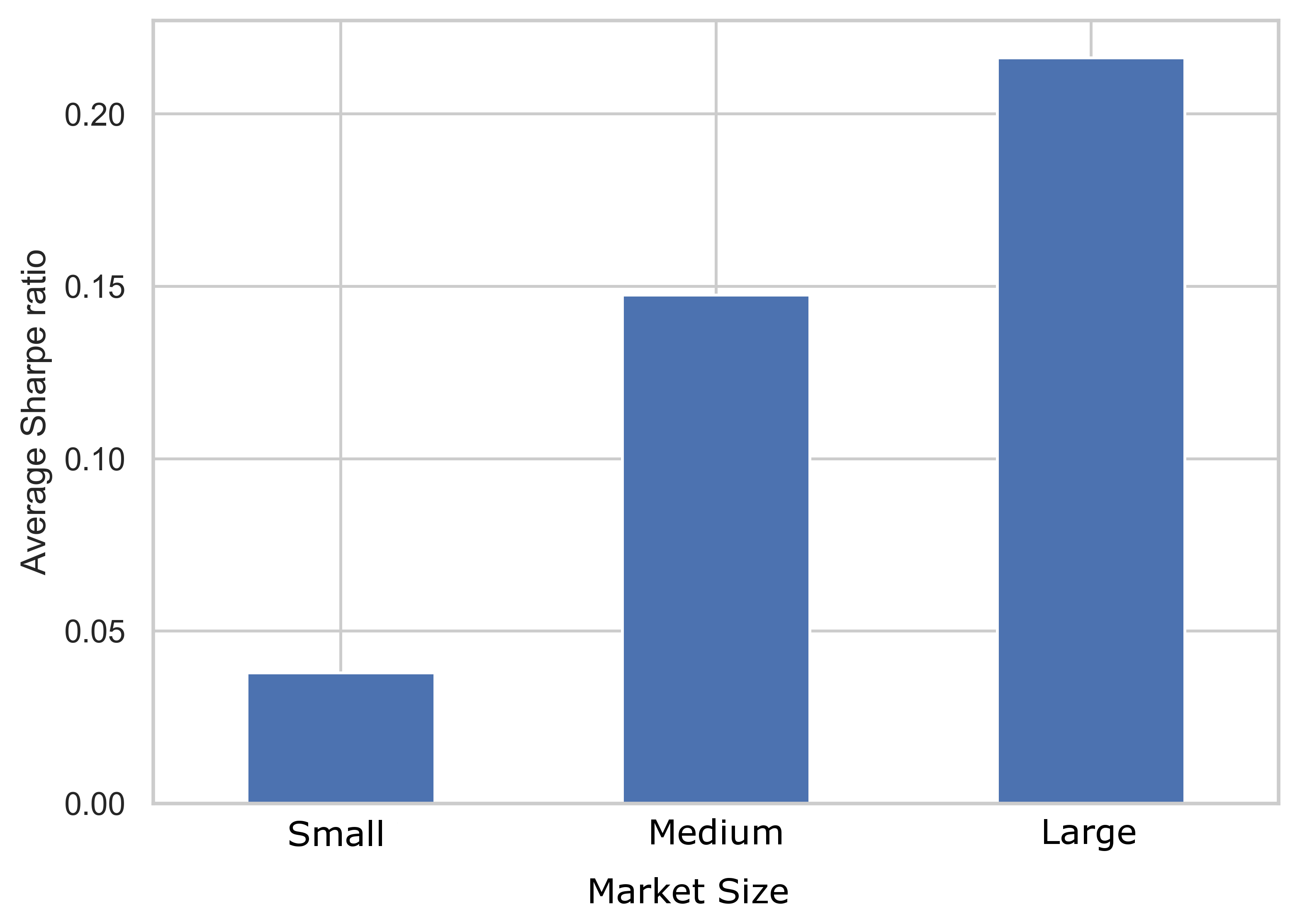}
		\caption{}
	\end{subfigure}
	\caption{(a) Average cumulative returns (\%) of SPX Index, UKX Index, KOSPI Index and SASEIDX Index contrasting QuantNet and No Transfer. Average Sharpe ratio difference between QuantNet versus No Transfer, aggregated by sample size (b) and number of assets per market (c) -- in both we have subtracted QuantNet SR from No Transfer SR to reduce cross-asset variance and baseline effect.}
	\label{fig:asset-sample-size-charts}
\end{figure}

Figure \ref{fig:barchart} outlines the average SR across the 58 markets, ordered by No Transfer strategy performance. In SR terms, QuantNet outperforms No Transfer in its top 5 markets and dominates the bottom 10 markets where No Transfer yields negative results, both in terms of SR and CR ratios. Finally, in 7 of the top 10 largest ones (RTY, SPX, KOSPI, etc.), QuantNet also outperforms No Transfer. Figure \ref{fig:asset-sample-size-charts}a presents cumulative returns charts in a set of large regional markets, such as United States S\&P 500 components (SPX Index), United Kingdom FTSE 100 (UKX Index), Korea Composite Index (KOSPI Index) and Saudi Arabia Tadawul All Shares (SASEIDX Index). Across regions, we observe a 2-10 times order of magnitude improvement in SRs and CRs by QuantNet, with similar benefits in Sortino ratios, Downside risks, and Skewness. Appendix \ref{in-depth} provides further analysis.

\paragraph{QuantNet Features} One of the key features of transfer learning is its ability to provide meaningful solutions in resource-constrained scenarios -- sample size, features, training budget, etc. With QuantNet this pattern persists; Figure \ref{fig:asset-sample-size-charts}b presents the average SR grouped based on market sample size in the training set. As transfer-learning would predict, we observe large gains to transfer in markets with tiny sample size  (1444-1823 samples or ~6-7 years) where fitting a model on only local market data yields poor performance. Further, gains from transfer generally decay as sample sizes increase. Interestingly, we find that medium-sized markets (2200-2576 samples or ~10 years of data) do not benefit from transfer, suggesting that there is room for improvement in the design of our transfer bottleneck $\omega$, an exciting avenue for future research. Another vital feature is coping with market size -- Figure \ref{fig:asset-sample-size-charts}c outlines QuantNet performance in terms of average SR. It demonstrates that the bigger the market, the better QuantNet will perform.


\section{Conclusion}

In this paper, we introduce \emph{QuantNet}: an architecture that learns market-agnostic trends and use these to learn superior market-specific trading strategies. QuantNet uses recent advances in transfer- and meta-learning, where market-specific parameters are free to specialize on the problem at hand, while market-agnostic parameters capture signals from all markets. QuantNet takes a step towards end-to-end global financial trading that can deliver superior market returns. In a few big regional markets, such as S\&P 500, FTSE 100, KOSPI and Saudi Arabia Tadawul All Shares, QuantNet showed 2-10 times improvement in SR and CR. QuantNet also generated positive and statistically significant alpha according to Fama-French 5 factors model (appendix \ref{fama-french}). An avenue of future research is to identify the functional form of a global transfer layer that can deliver strong performance also on markets where mixed transfer occurred, such as those with medium sample size. 


\section*{Broader Impact}

As this work proposes a machine learning system for financial trading, there are potential societal impacts. In principle, research that creates systems capable of matching expert trading performance provides a social good in that it can democratize financial trading. This is provided that the system can be trained (or is provided pre-trained) and can be deployed by individuals. Conversely, there are potential societal issues to technological advancements in financial trading; as system specialize and become increasingly complex, it is conceivable that they become less inclusive in their applicability, more opaque, and increase systemic risk.


We believe that while these potential benefits and risks apply in principle to our broad research direction, QuantNet itself is unlikely to have a significant societal impact. First, it is not meant for personal trading in its current form, and we provide no interface through which individuals or entities, in general, can make financial decisions. Second, QuantNet relies on well-known components, namely LSTMs, feed-forward networks, and stochastic gradient descent - all of which can be built with open-source software and trained on personal hardware. Hence we believe that QuantNet is widely accessible to the general public.



\bibliographystyle{plain}
\bibliography{bibthesis}

\newpage

\appendix

\section{Datasets} \label{datasets}

Table \ref{table:datasets} presents the datasets/markets used to empirically evaluate QuantNet. All the data was obtained via Bloomberg, with the description of each market/index and its constituents at \url{https://www.bloomberg.com}; for instance, SPX can be found by searching using the following link \url{https://www.bloomberg.com/quote/SPX:IND}. We tried to find a compromise between number of assets and sample size, hence for most markets we were unable to use the full list of constituents. We aimed to collect daily price data ranging from 03/01/2000 to 15/03/2019, but for most markets it starts roughly around 2010. Finally, due to restrictions from our Bloomberg license, we were unable to access data for some important equity markets, such as Italy and Russia. Full list with assets and respective exchange can be found at: \url{https://www.dropbox.com/s/eobhg2w8ithbgsp/AssetsExchangeList.xlsx?dl=0}

\begin{table}[h!]
	\tiny
	\caption{Markets used during our experiment. MEA - Middle East and Africa.}
	\label{table:datasets}
	\begin{tabular}{l|cccc|l|cccc}
		\hline
		\hline
		Region           & Index/Market & Country     & \# Samples & \# Assets & Region & Index/Market & Country      & \# Samples & \# Assets \\
		\hline
		Americas         & IBOV         & Brazil      & 3250        & 29               & Europe & HEX          & Finland      & 1882        & 65               \\
		Americas         & MERVAL       & Argentina   & 3055        & 11               & Europe & IBEX         & Spain        & 3499        & 23               \\
		Americas         & MEXBOL       & Mexico      & 3002        & 19               & Europe & ISEQ         & Ireland      & 2888        & 14               \\
		Americas         & RTY          & US          & 2356        & 554              & Europe & KFX          & Denmark      & 3345        & 15               \\
		Americas         & SPTSX        & Canada      & 3173        & 129              & Europe & OBX          & Norway       & 2812        & 17               \\
		Americas         & SPX          & US          & 3291        & 376              & Europe & OMX          & Sweden       & 3453        & 29               \\
		Asia and Pacific & AS51         & Australia   & 2363        & 91               & Europe & PX           & Czechia      & 3374        & 5                \\
		Asia and Pacific & FBMKLCI      & Malaysia    & 3131        & 23               & Europe & SBITOP       & Slovenia     & 2995        & 6                \\
		Asia and Pacific & HSI          & China       & 2599        & 37               & Europe & SMI          & Switzerland  & 3948        & 19               \\
		Asia and Pacific & JCI          & Indonesia   & 2007        & 44               & Europe & SOFIX        & Bulgaria     & 1833        & 5                \\
		Asia and Pacific & KOSPI        & South Korea & 3041        & 297              & Europe & UKX          & UK           & 3664        & 75               \\
		Asia and Pacific & KSE100       & Pakistan    & 2036        & 41               & Europe & VILSE        & Lithuania    & 2765        & 5                \\
		Asia and Pacific & NIFTY        & India       & 3066        & 38               & Europe & WIG20        & Poland       & 3449        & 8                \\
		Asia and Pacific & NKY          & Japan       & 3504        & 186              & Europe & XU100        & Turkey       & 2545        & 76               \\
		Asia and Pacific & NZSE50FG     & New Zealand & 3258        & 21               & MEA    & DFMGI        & UAE          & 2184        & 11               \\
		Asia and Pacific & PCOMP        & Philippines & 3013        & 16               & MEA    & DSM          & Qatar        & 2326        & 16               \\
		Asia and Pacific & SHSZ300      & China       & 2881        & 18               & MEA    & EGX30        & Egypt        & 1790        & 22               \\
		Asia and Pacific & STI          & Singapore   & 2707        & 27               & MEA    & FTN098       & Namibia      & 1727        & 16               \\
		Asia and Pacific & TWSE         & Taiwan      & 3910        & 227              & MEA    & JOSMGNFF     & Jordan       & 2287        & 15               \\
		Europe           & AEX          & Netherlands & 4083        & 17               & MEA    & KNSMIDX      & Kenya        & 1969        & 14               \\
		Europe           & ASE          & Greece      & 2944        & 51               & MEA    & KWSEPM       & Kuwait       & 2785        & 11               \\
		Europe           & ATX          & Austria     & 3511        & 13               & MEA    & MOSENEW      & Morocco      & 2068        & 27               \\
		Europe           & BEL20        & Belgium     & 3870        & 14               & MEA    & MSM30        & Oman         & 2069        & 24               \\
		Europe           & BUX          & Hungary     & 3753        & 8                & MEA    & NGSE30       & Nigeria      & 1761        & 25               \\
		Europe           & BVLX         & Portugal    & 3269        & 17               & MEA    & PASISI       & Palestine    & 1447        & 5                \\
		Europe           & CAC          & France      & 3591        & 36               & MEA    & SASEIDX      & Saudi Arabia & 1742        & 71               \\
		Europe           & CRO          & Croatia     & 1975        & 13               & MEA    & SEMDEX       & Mauritius    & 2430        & 5                \\
		Europe           & CYSMMAPA     & Cyprus      & 2056        & 42               & MEA    & TA-35        & Israel       & 2677        & 23               \\
		Europe           & DAX          & Germany     & 3616        & 27               & MEA    & TOP40        & South Africa & 2848        & 34              \\
		\hline
		\hline
	\end{tabular}
\end{table}

\section{Evaluation} \label{benchmarking}

\paragraph{Baselines}We compared QuantNet with four other traditional and widely adopted and researched trading strategies. Below we briefly expose each one of them as well as provide some key references:
\begin{itemize}
    \item \textbf{Buy and hold}: this strategy simply purchase a unit of stock and hold it, that is, $\boldsymbol{s}^{\text{BaH}}_{t} := \boldsymbol{1}$ for all assets in a market. Active trading strategies are supposed to beat this passive strategy, but in some periods just holding a S\&P 500 portfolio passively outperform many active managed funds \cite{elton2019passive,dichtl2020investing}.
    
    \item \textbf{Risk parity}: this approach trade assets in a certain market such that they contribute as equally as possible to the portfolio overall volatility. A simple approach used is to compute signals per asset as
    \begin{equation}
        s^{\text{RP}}_{t, j} := \frac{ \frac{1}{\sigma^{j}_{t:t-252}} }{ \frac{1}{\sum_{j=1}^n \sigma^{j}_{t:t-252}} }
    \end{equation}
    with $\sigma^{j}_{t:t-252}$ as the rolling 252 days ($\approx$ 1 year) volatility of asset $j$. Interest in the risk parity approach has increased since the late 2000s financial crisis as the risk parity approach fared better than traditionally constructed portfolios \cite{choueifaty2008toward, martellini2008toward, du2016volatility}.
    
    \item \textbf{Time series momentum}: this strategy, also called trend momentum or trend-following, suggest going long in assets which have had recent positive returns and short assets which have had recent negative returns. It is possibly one of the most adopted and researched strategy in finance \cite{moskowitz2012time,daniel2016momentum,baltzer2019trades}. For a given asset, the signal is computed as
    \begin{equation}
        s^{\text{TSMOM}}_{t, j} := \mu^{j}_{t:t-252}
    \end{equation}
    with 252 days ($\approx$ 12 months, $\approx$ 1 year) the typical lookback period to compute the average return $\mu^{j}_{t:t-252}$ of asset $j$.
    
    \item \textbf{Cross-sectional momentum}: the cross-sectional momentum strategy as defined by is a long-short zero-cost portfolio that consists of securities with the best and worst relative performance over a lookback period \cite{jegadeesh1993returns, baz2015dissecting, feng2020taming}. It works similarly as time series momentum, with the addition of screening weakly performing and underperfoming assets. For a given market, the signal can be computed as
    \begin{equation}
        s^{\text{CSMOM}}_{t, j} := 
        \begin{cases}
        \mu^{j}_{t:t-252}, \quad \text{if} \mu^{j}_{t:t-252} > Q_{1-q}(\mu^{1}_{t:t-252}, ..., \mu^{n}_{t:t-252}) \\
        -\mu^{j}_{t:t-252}, \quad \text{if} \mu^{j}_{t:t-252} < Q_{q}(\mu^{1}_{t:t-252}, ..., \mu^{n}_{t:t-252}) \\
        0, \quad \text{otherwise}
        \end{cases}
    \end{equation}
    with $Q_{q}(\mu^{1}_{t:t-252}, ..., \mu^{n}_{t:t-252})$ representing the q-th quantile of the assets average returns. A signal for going long (short) is produced if the asset $j$ is at the top (bottom) quantile of the distribution. In our experiments we used the typical value of $q=0.33$.
    
\end{itemize}

\paragraph{Hyperparameters}Table \ref{table:hyperparameters} outlines the settings for QuantNet and No Transfer strategies. Since running an ehxaustive search is computationally prohibitive, we opted to use random search as our hyperparameter optimization strategy \cite{bergstra2012random}. We randomly sampled a total of 200 values in between those ranges, giving larger bounds for configurations with less hyperparameters (No Transfer linear and QuantNet Linear-Linear). After selecting the best hyperparameters, we applied them in a holdout-set consisting of the last 752 observations of each time series (around 3 years). The metrics and statistics in this set are reported in our results section. After a few warm-up runs, we opted to use 2000 training steps as a good balance between computational time and convergence. We trained the different models using the stochastic gradient descent optimizer AMSgrad \cite{reddi2019convergence}, a variant of the now ubiquitously used Adam algorithm.

\begin{table}[h!]
	\centering
	\scriptsize
	\caption{No Transfer and QuantNet hyperparameters and configurations investigated.}
	\label{table:hyperparameters}
	\begin{tabular}{l|cc|cccc}
		\hline
		\hline
		Hyper-          & \multicolumn{2}{l}{No Transfer} & \multicolumn{4}{l}{\begin{tabular}[c]{@{}l@{}}Quantnet (Encoder/Decoder-Transfer Layer)\end{tabular}} \\
		parameter       & Linear         & LSTM           & Linear-Linear              & Linear-LSTM              & LSTM-Linear              & LSTM-LSTM              \\
		\hline
		Batch size ($L$)     & 16-128         & 16-128         & 16-128                     & 16-96                    & 16-96                    & 16-96                  \\
		Sequence length ($p$) & 21-504         & 21-504         & 21-504                     & 21-252                   & 21-252                   & 21-252                 \\
		Learning rate   & 0.0001-0.1     & 0.0001-0.1     & 0.0001-0.1                 & 0.0001-0.5               & 0.0001-0.5               & 0.0001-0.5             \\
		E/D \# layers   &                & 1-2            &                            &                          & 1-2                      & 1-2                    \\
		E/D dropout     &                & 0.1-0.9        &                            &                          & 0.1-0.9                  & 0.1-0.9                \\
		TL \# layers    &                &                &                            & 1-2                      &                          & 1-2                    \\
		TL dropout      &                &                &                            & 0.1-0.9                  &                          & 0.1-0.9                \\
		TL dimension ($N$)    & \multicolumn{6}{c}{10, 25, 50, 100}                                 \\
		Training steps    & \multicolumn{6}{c}{2000}                                 \\
		\hline
		\hline
	\end{tabular}
\end{table}

\paragraph{Financial metrics}We have used 3 Month London Interbank Offered Rate in US Dollar as the reference rate to compute excess returns. Most of the results focus on Sharpe ratios, but in many occasions we have also reported Calmar ratios, Annualized Returns and Volatility, Downside risk, Sortino ratios, Skewness and Maximum drawdowns \cite{young1991calmar,sharpe1994sharpe,eling2007does,rollinger2013sortino}. 

\section{LSTMs and QuantNet's Architecture} \label{architecture}



Given as inputs a sequence of returns from a history $R_{m:t} = (\boldsymbol{r}_{t-m}^i, \ldots, \boldsymbol{r}_t^i)$ of market $i$, below we outline QuantNet's input to trading signal (output) mapping, considering the LSTM and Linear models \cite{hochreiter1997long,gers1999learning,flennerhag2018breaking} defined by the gating mechanisms:
\begin{equation}
\boldsymbol{e}_{t}^{i} = LSTM(\boldsymbol{r}_{t-1}^{i}, \boldsymbol{e}_{t-1}^{i}) = 
\begin{cases}
\boldsymbol{u}_{t}^{s \in \{p, f, o, g\}} = W^{(s)}_{\boldsymbol{e}^i} \boldsymbol{r}_{t-1} + V^{(s)}_{\boldsymbol{e}^i} \boldsymbol{e}_{t-1}^{i} + \boldsymbol{b}^{(s)}_{\boldsymbol{e}^i} \\
\boldsymbol{c}_{t}^{\boldsymbol{e}^i} = \sigma(\boldsymbol{u}_{t}^{f}) \odot \boldsymbol{c}_{t-1}^{\boldsymbol{e}^i} + \sigma(\boldsymbol{u}_{t}^{p}) \odot \tanh(\boldsymbol{u}_{t}^{g}) \\
\boldsymbol{e}_{t}^{i} = \sigma(\boldsymbol{u}_{t}^{o}) \odot \tanh(\boldsymbol{c}_{t}^{\boldsymbol{e}^i})
\end{cases}
\label{ec-eq}
\end{equation}

\begin{equation}
\boldsymbol{z}_{t}^{i} = \omega(\boldsymbol{e}_{t}^{i}) = Z \boldsymbol{e}_{t}^{i} + \boldsymbol{b}_Z \label{tl-params} \\
\end{equation}

\begin{equation}
\boldsymbol{d}_{t}^{i} = LSTM(\boldsymbol{z}_{t}^{i}, \boldsymbol{d}_{t-1}^{i}) = 
\begin{cases}
\boldsymbol{v}_{t}^{s \in \{p, f, o, g\}} = W^{(s)}_{\boldsymbol{d}^i} \boldsymbol{z}_{t}^{i} + V^{(s)}_{\boldsymbol{d}^i} \boldsymbol{d}_{t-1}^{i} + \boldsymbol{b}^{(s)}_{\boldsymbol{d}^i} \\
\boldsymbol{c}_{t}^{\boldsymbol{d}^i} = \sigma(\boldsymbol{v}_{t}^{f}) \odot \boldsymbol{c}_{t-1}^{\boldsymbol{d}^i} + \sigma(\boldsymbol{v}_{t}^{p}) \odot \tanh(\boldsymbol{v}_{t}^{g}) \\
\boldsymbol{d}_{t}^{i} = \sigma(\boldsymbol{v}_{t}^{o}) \odot \tanh(\boldsymbol{c}_{t}^{\boldsymbol{d}^i})
\end{cases}
\label{dc-eq}
\end{equation}
\begin{equation}
\boldsymbol{s}_{t}^{i} = \tanh(W^i \boldsymbol{d}_{t}^{i} + \boldsymbol{b}^i) \label{trad-signal}
\end{equation}

where $\sigma$ represents the sigmoid activation function, and $\boldsymbol{u}_{t}^{s \in \{p, f, o, g\}}$ and $\boldsymbol{v}_{t}^{s \in \{p, f, o, g\}}$ linear transformations. The remaining components are the encoder cell $\boldsymbol{c}_{t}^{\boldsymbol{e}^i}$ and hidden state $\boldsymbol{e}_{t}^{i}$ (eq. \ref{ec-eq}); Linear transfer layer mapping $\boldsymbol{z}_{t}^{i}$ (eq. \ref{tl-params}); decoder cell $\boldsymbol{c}_{t}^{\boldsymbol{d}^i}$ and hidden state $\boldsymbol{d}_{t}^{i}$ (eq. \ref{dc-eq}); and final long-short trading signal $\boldsymbol{s}_{t}^{i} \in [-1, 1]$  (eq. \ref{trad-signal}). In QuantNet, we interleave market specific and market agnostic parameters in the model. Each market is therefore associated with specific parameters $W^{(s)}_{\boldsymbol{e}^i}, W^{(s)}_{\boldsymbol{d}^i}, V^{(s)}_{\boldsymbol{e}^i}, W^{(s)}_{\boldsymbol{d}^i}, \boldsymbol{b}^{(s)}_{\boldsymbol{e}^i}, \boldsymbol{b}^{(s)}_{\boldsymbol{d}^i}, W^i,  \boldsymbol{b}^i$, while all markets share parameters $Z$ and $\boldsymbol{b}_Z$ (eq. \ref{tl-params}). 

\section{Literature Review} \label{literature}

In this section we aim to provide a general view of the different subareas inside transfer learning (rather than a thorough review about the whole area). With this information, our goal is to frame the current contributions in finance over these subareas, situate our paper contribution, as well as highlight outstanding gaps. Nonetheless, the reader interested in a thorough presentation about transfer learning should refer to these key references \cite{pan2009survey,Book:GoodfellowDeep:2016,Paper:SurveyTransfer:2019}

\subsection{Transfer Learning: definition}

We start by providing a definition of transfer learning, mirroring notation and discussions in \cite{pan2009survey,Paper:RuderTransfer:2019,Paper:SurveyTransfer:2019}. A typical transfer learning problem presume the existence of a domain and a task. Mathematically, a domain $\mathcal{D}$ comprises a feature space $X \in \mathcal{X}$ and a probability measure $\mathbb{P}$ over $X$, where $\mathbf{x}_i = \{x_1, ..., x_J\}$ is a realization of $X$. As an example for trading strategies, $\mathcal{X}$ can be the space of all technical indicators, $X$ a specific indicator (e.g. book-to-market ratio), and $\mathbf{x}_i$ a random sample of indicators taken from $X$. Given a domain $\mathcal{D} = \{\mathcal{X}, \mathbb{P}(X) \}$ and a supervised learning setting, a task $\mathcal{T}$ consists of a label space $Y \in \mathcal{Y}$, and a conditional probability distribution $\mathbb{P}(Y|X)$\footnote{A more generic definition, that works for unsupervised and reinforcement learning, demands that along with every task $\mathcal{T}_i$ we have an objective function $f_i$.}. Typically in trading strategies, $\mathcal{Y}$ can represent the next quarter earnings, and $\mathbb{P}(Y|X)$ is learned from the training data $(\mathbf{x}_i, y_i)$.

\begin{figure}[h!]
	\centering
	\includegraphics[width=0.85\textwidth]{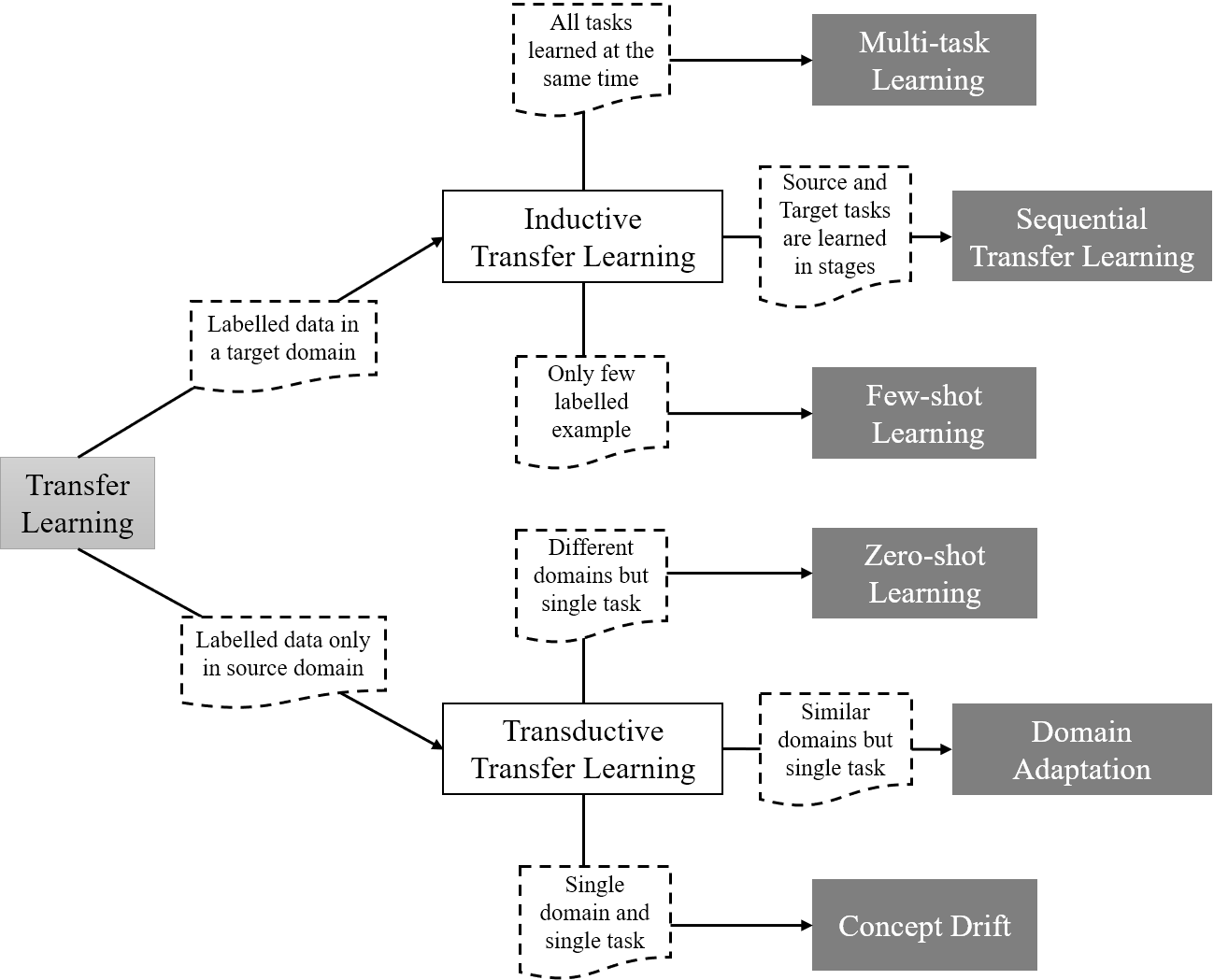}
	\caption{Taxonomy of transfer learning sub-paradigms.}
	\label{fig:TransferLearningAreas}
\end{figure}

The domain $\mathcal{D}$ and task $\mathcal{T}$ are further split in two subgroups: source domains $\mathcal{D}_S$ and corresponding tasks $\mathcal{T}_S$, as well as target domain $\mathcal{D}_T$ and target task $\mathcal{T}_T$. Therefore, the objective of transfer learning is to learn the target conditional probability distribution $\mathbb{P}_T(Y_T|X_T)$ in $\mathcal{D}_T$ with information gained from $\mathcal{D}_S$ and $\mathcal{T}_S$. Usually, either a limited number of labelled target examples or a large number of unlabelled target examples are assumed to be available. The way this learning is performed across the tasks, the amount of labelled information as well as inequalities between $\mathcal{D}_S$ and $\mathcal{D}_T$, and $\mathcal{T}_S$ and $\mathcal{T}_T$ give rise to different forms of transfer learning. Figure \ref{fig:TransferLearningAreas} presents these different scenarios\footnote{We should note that there are other possibilities, but they fall into Unsupervised or Reinforcement transfer learning. These other paradigms fall outside the scope of this work, which is mostly interested in (Semi-) Supervised transfer learning.}. 

In what follows we analyse each sub-paradigm of Figure \ref{fig:TransferLearningAreas}.

\subsection{Transfer Learning: sub-paradigms}

\textbf{Inductive Transfer Learning}: it refers to the cases where labelled data is available in the target domain; in another sense, we have the typical Supervised learning scenario across the different domains and tasks. In inductive transfer methods, the target-task inductive bias is chosen or adjusted based on the source-task knowledge. The way this is done varies depending on which inductive learning algorithm is used to learn the source and target tasks \cite{Paper:LisaTorrey:2010}. The main variations occur on how this learning is performed: simultaneously across source and target tasks (Multi-Task); sequentially by sampling source tasks, and updating the target task model (Sequential Transfer); and with the constraints of using only few labelled examples (Few-shot). We outline each variation:

\begin{itemize}
	\item Multi-task Learning \cite{caruana1997multitask,Paper:SurveyMTL:2017}: is an approach to inductive transfer that improves generalization by learning tasks \emph{in parallel} while using a shared representation; hence, $\mathbb{P}_T(Y_T|X_T)$ and $\mathbb{P}_S(Y_S|X_S)$ are intertwined, with the update in the target task affecting the behaviour of the domain tasks, and vice-versa. In practice, the learned model architecture and parameters are fully-shared across domain and target tasks -- inputs, weights or coefficients, transfer functions, and objective function. In finance, this mode of learning has been first used for stock selection \cite{ghosn1997multi}; lately, it has been applied for day trading \cite{bitvai2015day} and yield curves \cite{nunes2019comparison}.
	
	\item Sequential Transfer Learning \cite{Paper:RuderTransfer:2019}: is an approach to inductive transfer that improves generalization by learning tasks \emph{in sequence} while using a shared representation to a certain extent; therefore, $\mathbb{P}_T(Y_T|X_T)$ and $\mathbb{P}_S(Y_S|X_S)$ are not completely intertwined, but the update in the target task impacts the behaviour of the domain tasks, and vice-versa. In practice, the learned model architecture and parameters are partially-shared across domain and target tasks -- often weights, transfer functions, and sometimes the objective function. By not having to share the same inputs and other parts of the architecture, this mode of learning can be applied across different domains and make the learned model easier to reused in future tasks. In the context of financial applications, it has been mainly applied for sentiment analysis: One of such applications is FinBERT \cite{araci2019finbert}, a variation of BERT \cite{Paper:BERT:2018} specialized to financial sentiment analysis; it has obtained state-of-the-art results on FiQA sentiment scoring and Financial PhraseBank benchmaks. In \cite{hiew2019bert} provide a similar application but feeding the sentiment analysis index generated by BERT in a LSTM-based trading strategy to predict stock returns.
	
	\item Few-shot Learning \cite{Paper:FeiFeiOne:2006,Book:GoodfellowDeep:2016,wang2019generalizing}: it is an extreme form of inductive learning, with very few examples (sometimes only one) being used to learn the target task model. This works to the extent that the factors of variation corresponding to these invariances have been cleanly separated from the other factors, in the learned representation space, and that we have somehow learned which factors do and do not matter when discriminating objects of certain categories. During the transfer learning stage, only a few labeled examples are needed to infer the label of many possible test examples that all cluster around the same point in representation space. So far we were unable to find any application in finance that covers this paradigm. However, we believe that such a mode of learning can be applied for fraud detection, stock price forecasting that have recently undergone initial public offering, or any other situation where limited amount of data is present about the target task.
\end{itemize}

\textbf{Transductive Transfer Learning}: it refers to the cases where labelled data is only available in the source domain, although our objective is still to solve the target task; hence, we have a situation that is somewhat similar to what is known as Semi-supervised learning. What makes the transductive transfer methods feasible is the fact that the source and target tasks are the same, although the domains can be different. For example, consider the task of sentiment analysis, which consists of determining whether a comment expresses positive or negative sentiment. Comments posted on the web come from many categories. A transductive sentiment predictor trained on customer reviews of media content, such as books, videos and music, can be later used to analyze comments about consumer electronics, such as televisions or smartphones. There are three main forms of transductive transfer learning: Domain Adaptation, Concept Drift and Zero-shot learning. Each form is presented below: 

\begin{itemize}
	\item Domain Adaptation \cite{kouw2019review}: in this case the tasks remains the same between each setting, but the domains as well as the input distribution are usually slightly different; therefore $\mathcal{X}_S \approx \mathcal{X}_T$ or $\mathbb{P}_T(X_T) \approx \mathbb{P}_S(X_S)$. The previous example of the sentiment predictor is a typical case, where the domains and the input distribution is somewhat different (books, videos and music reviews transferring to consumer electronics). We can presume that there is an underlying mapping that matches a certain statement to positive, neutral or negative sentiment, and what makes the problem harder to solve is the fact that the vocabulary and context vary between domains. Surprisingly simple unsupervised pretraining has been found to be very successful for sentiment analysis with domain adaptation \cite{glorot2011domain}. Similarly to Few-shot learning, this particular subarea of transfer learning has received less attention from the finance community, since most of the sentiment analysis and similar applications are handled using labelled data. 
	
	\item Concept Drift \cite{vzliobaite2016overview, escovedo2018detecta}: in this case the tasks and domains remains the same across settings, but the input distribution can gradually or abruptly change between them; therefore $\mathbb{P}_T(X_T) \neq \mathbb{P}_S(X_S)$. Often concept drift modelling and detection focus on continuous data streams, such as time series, text messages, videos, that is, data with a temporal dimension or indexation. Using the previous example, we would be concerned with changing views about a specific film: reviews that were otherwise extensively positive, gradually become negative due to changes in audience's view about how certain characters were portrayed, how the topic was approached, etc. This particular subarea has received substantial attention from the finance community: it has been used to discover relations between portfolio selection factors and stock returns \cite{hu2015concept}; price forecasting \cite{liu2019meta}; and fraud detection \cite{somasundaram2019parallel}.
	
	\item Zero-shot Learning \cite{Paper:SocherZero:2013,wang2019survey}: is a form of transductive transfer learning, where the domains and input distributions are different, and yet learning can be achieved by finding a suitable representation; hence $\mathcal{X}_S \neq \mathcal{X}_T$ and $\mathbb{P}_T(X_T) \neq \mathbb{P}_S(X_S)$. Following the previous example, if we have a database with thorough reviews about road bicycles, such as describing their frame, suspension, drivetrain, etc. it would be possible to learn in principle what constitutes a good or bad bicycle. Zero-shot learning would attempt to tap into this knowledge, and transfer it to a new bicycle that we do not have reviews but use it's design, 3d images, other descriptions, etc. to come up with an expected score, just based on users' opinions about the product. In this case, the task is the same (deciding the expected review of bicycle), but the domains are radically different (textual description versus an image). Similar to Few-shot learning, we were unable to identify any piece of research from the finance community.  
\end{itemize}

Also, there are four different approaches where the transference of knowledge from a task to another can be realized: instance, feature, parameters, and relational-knowledge. Table \ref{tab:TransferLearningApproaches} presents a brief description, applications of each to the financial domain, and other key references. Undoubtedly, for financial applications parameter-transfer is the preferred option, followed by instance-transfer and feature-transfer. Such approaches are mainly used for sentiment analysis, fraud detection, and forecasting, areas that have been widely researched using more traditional techniques. Conversely, we were unable to find research for relational-knowledge. Despite that, we believe that researchers working on financial networks, peer-to-peer lending, etc. can benefit from methods in the relational-knowledge transfer approach.

\begin{table}[h!]
	\centering
	\caption{Approaches and applications of transfer learning across Finance and general domains.}
	\small
	\begin{tabular}{c|p{4cm}|p{4cm}|c}
		\hline
		\hline
		\multirow{2}{*}{Approach} & \multirow{2}{*}{Brief Description} & \multirow{2}{*}{Financial Applications} & Other \\
		& &  & References \\
		\hline
		\hline
		\multirow{3}{*}{Instance} & Re-weighting labelled data in the source domain for use in the target domain & News-rich to news-poor stocks \cite{li2018stock}; mitigating class imbalance in credit scoring \cite{li2018transfer,voumardtransfer2019} & \cite{jiang2019deep,qu2019learning}\\
		\hline
		\multirow{3}{*}{Feature} & Find a suitable feature mapping to approximate the source domain to the target domain & Sentiment feature space \cite{li2018stock}; Portfolio selection factors \cite{hu2015concept} & \cite{zhuang2015supervised,yang2018glomo} \\
		\hline
		\multirow{4}{*}{Parameter} & Learn shareable parameters or priors between the source and target tasks models & BERT specialized to financial sentiment analysis \cite{araci2019finbert,hiew2019bert}; Stock selection, forecasting \cite{ghosn1997multi,bitvai2015day}; yield curve forecasting \cite{nunes2019comparison}  &  \cite{Paper:BERT:2018,chen2019med3d} \\
		\hline
		\multirow{3}{*}{Relational-knowledge} & Learn a logical relationship or rules in the source domain and transfer it to the target domain & & \cite{Park_2019_CVPR,Wang:2016:RKT:3016100.3016198} \\
		\hline
		\hline
	\end{tabular}
	\label{tab:TransferLearningApproaches}
\end{table}

Using this taxonomy, \emph{QuantNet can be classified as part of Sequential Transfer Learning, using a Parameter-transfer approach}. In this sense, we aim to learn the target task model by sharing and updating the architecture's weights and activation functions across the tasks. Since each task has different number of inputs/outputs, this component is task-specific. All of these details are better outlined in the next section.
\newpage
\section{In-depth Comparison: QuantNet and No Transfer Linear} \label{in-depth}

\begin{figure}[h!]
	\centering
	\begin{subfigure}{.5\textwidth}
		\centering
		\includegraphics[width=\linewidth]{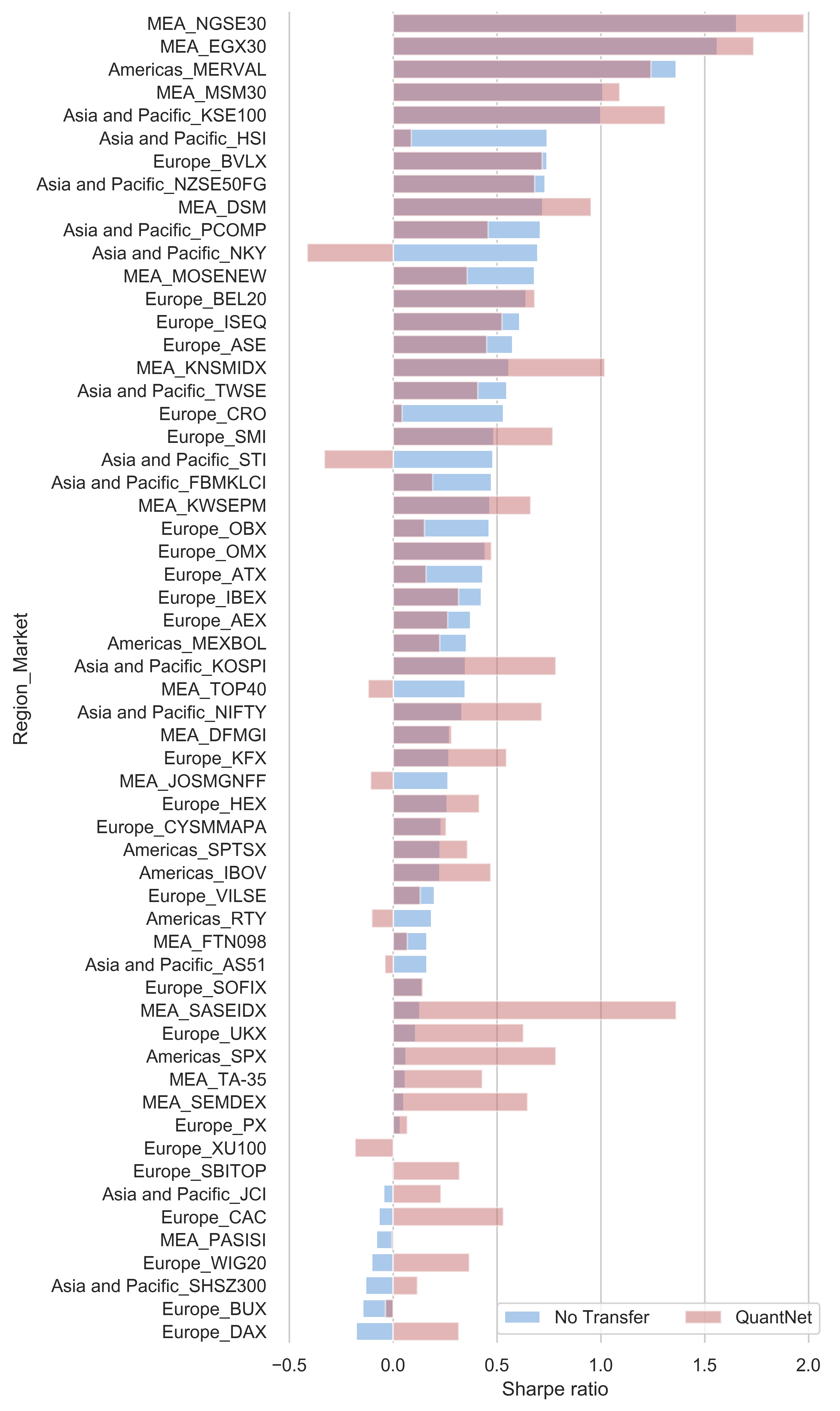}
		\caption{}
	\end{subfigure}%
	\begin{subfigure}{.5\textwidth}
		\centering
		\includegraphics[width=\linewidth]{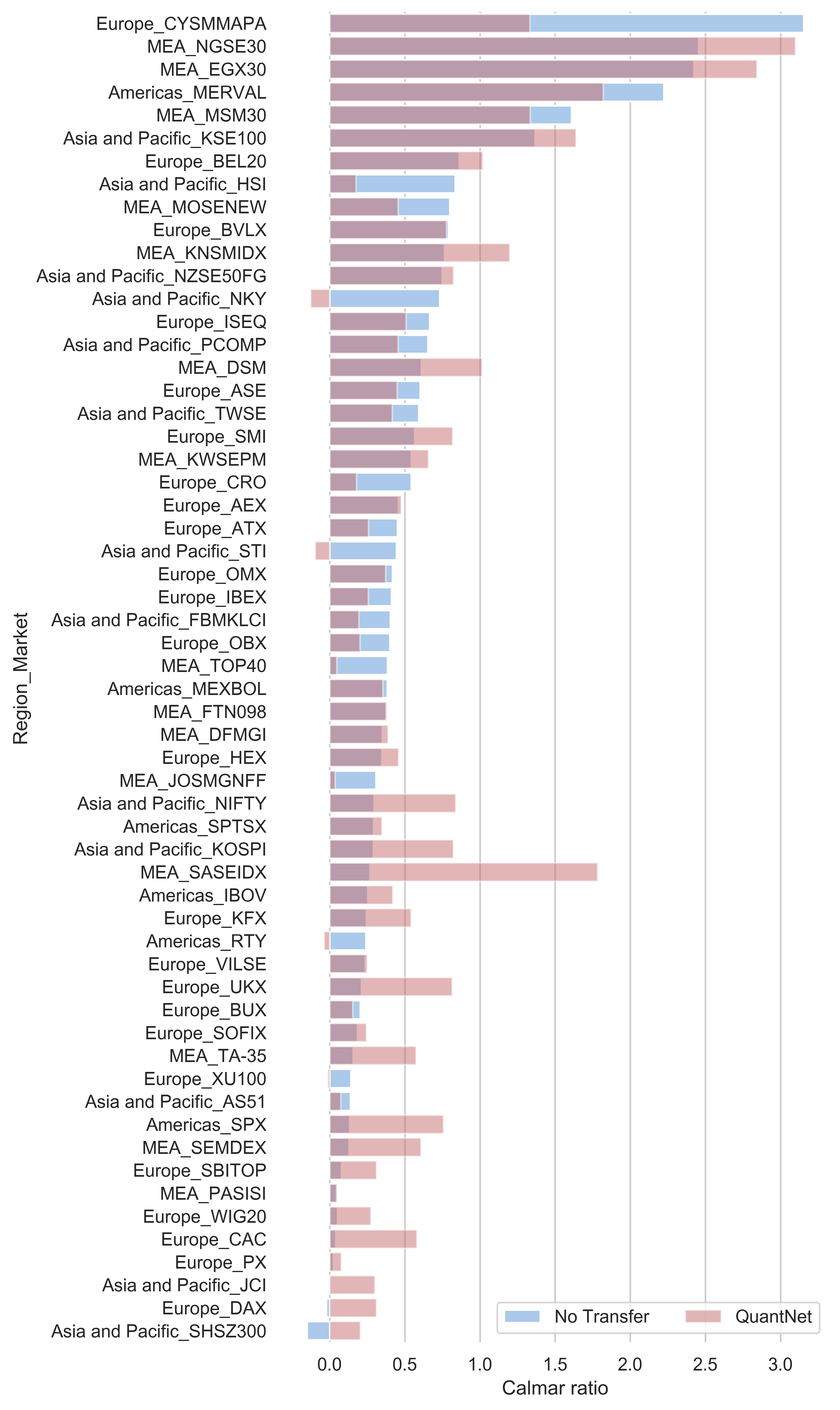}
		\caption{}
	\end{subfigure}
	\caption{Average Sharpe (a) and Calmar (b) ratios of QuantNet and No Transfer across 58 markets.}
	\label{fig:market-column-charts}
\end{figure}

\paragraph{Market-level analysis} Figures \ref{fig:market-column-charts}a and \ref{fig:market-column-charts}b outline the average SR and CR across the 58 markets, ordered by No Transfer strategy performance. In SR terms, QuantNet outperformed No Transfer in its top 5 markets, and dominates the bottom 10 markets both in SR and CR terms. Finally, in 7 of the top 10 largest ones (RTY, SPX, KOSPI, etc., see Table \ref{table:datasets}), QuantNet has also outperformed No Transfer. 

\begin{figure}[h!]
	\centering
	\includegraphics[width=\textwidth]{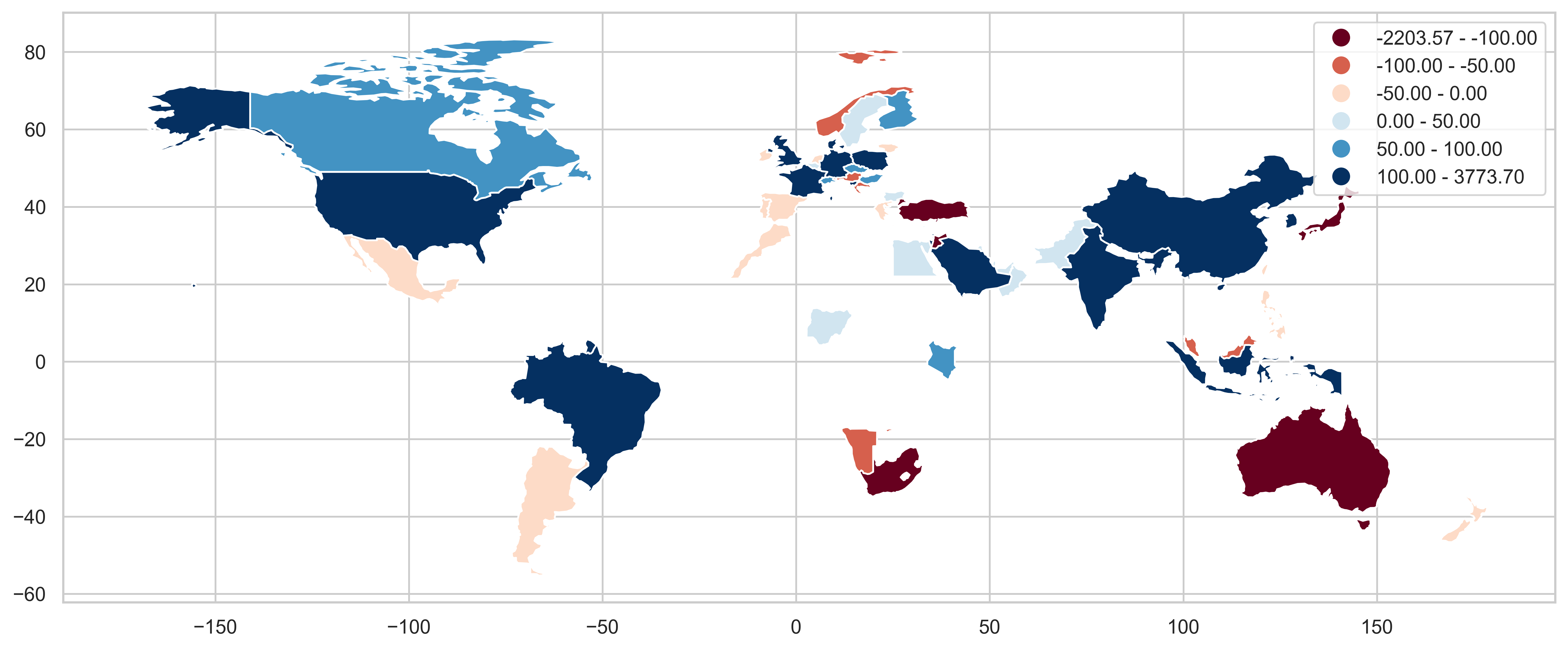}
	\caption{World map of average relative (\%) Sharpe ratio difference between QuantNet versus No Transfer. For visualisation purposes we have averaged the metric for US, China and Israel/Palestine.}
	\label{fig:map-spreadrel-sharpe-ratio}
\end{figure}

Figure \ref{fig:map-spreadrel-sharpe-ratio} maps every market to a country, and displays the relative outperformance (\%) of QuantNet in relation to No Transfer in SR values. In the Americas, apart from Mexico and Argentina, Brazil, US (on average) and Canada, QuantNet has produced better results than No Transfer. Similarly, the core of Europe (Germany, United Kingdom and France), and India and China, QuantNet has produced superior SRs than No Transfer, with markets like Japan, Australia, New Zealand, and South Africa representing the reverse.

\begin{figure}[h!]
	\centering
	\begin{subfigure}{.5\textwidth}
		\centering
		\includegraphics[width=\linewidth]{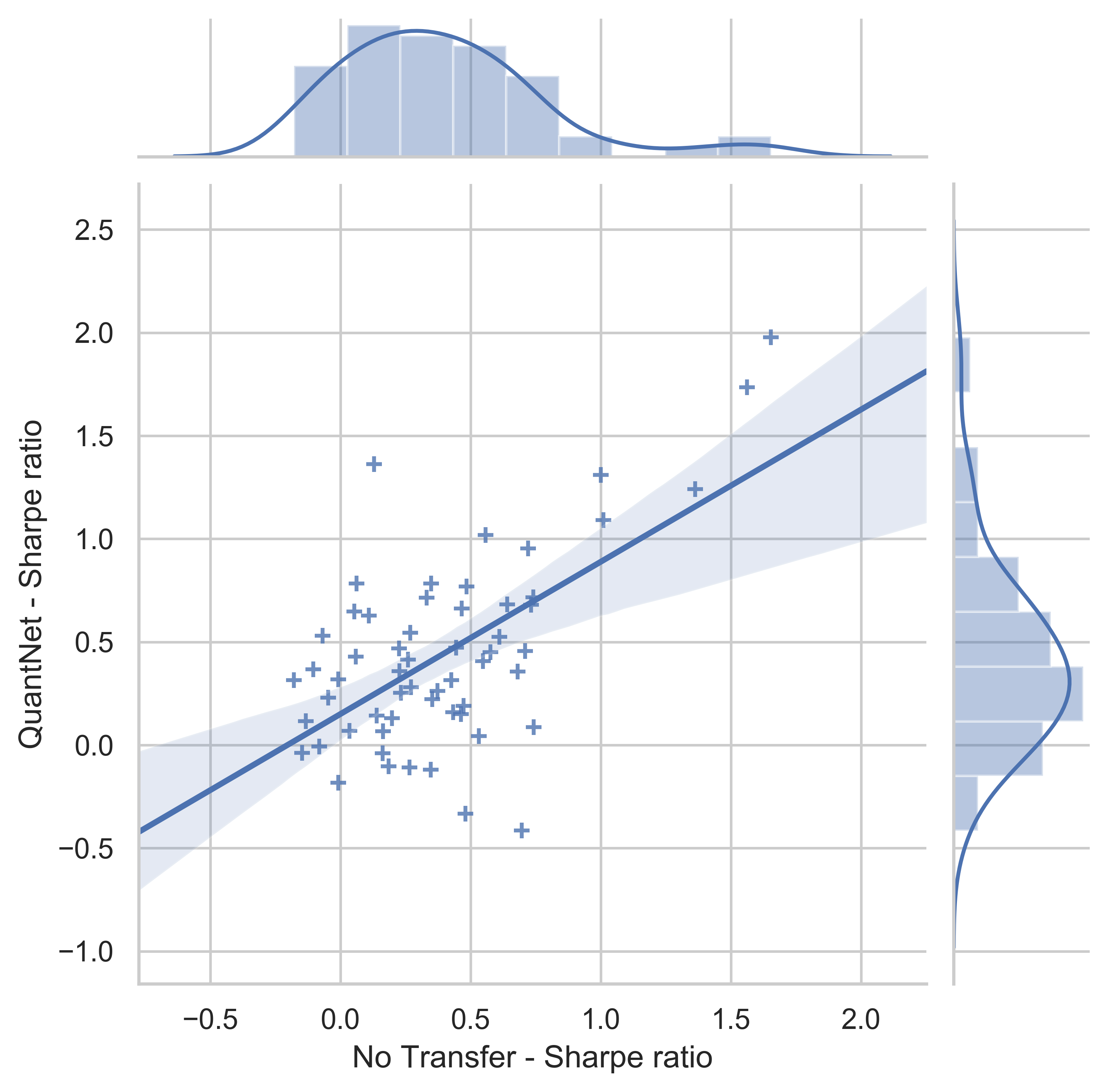}
		\caption{}
	\end{subfigure}%
	\begin{subfigure}{.5\textwidth}
		\centering
		\includegraphics[width=\linewidth]{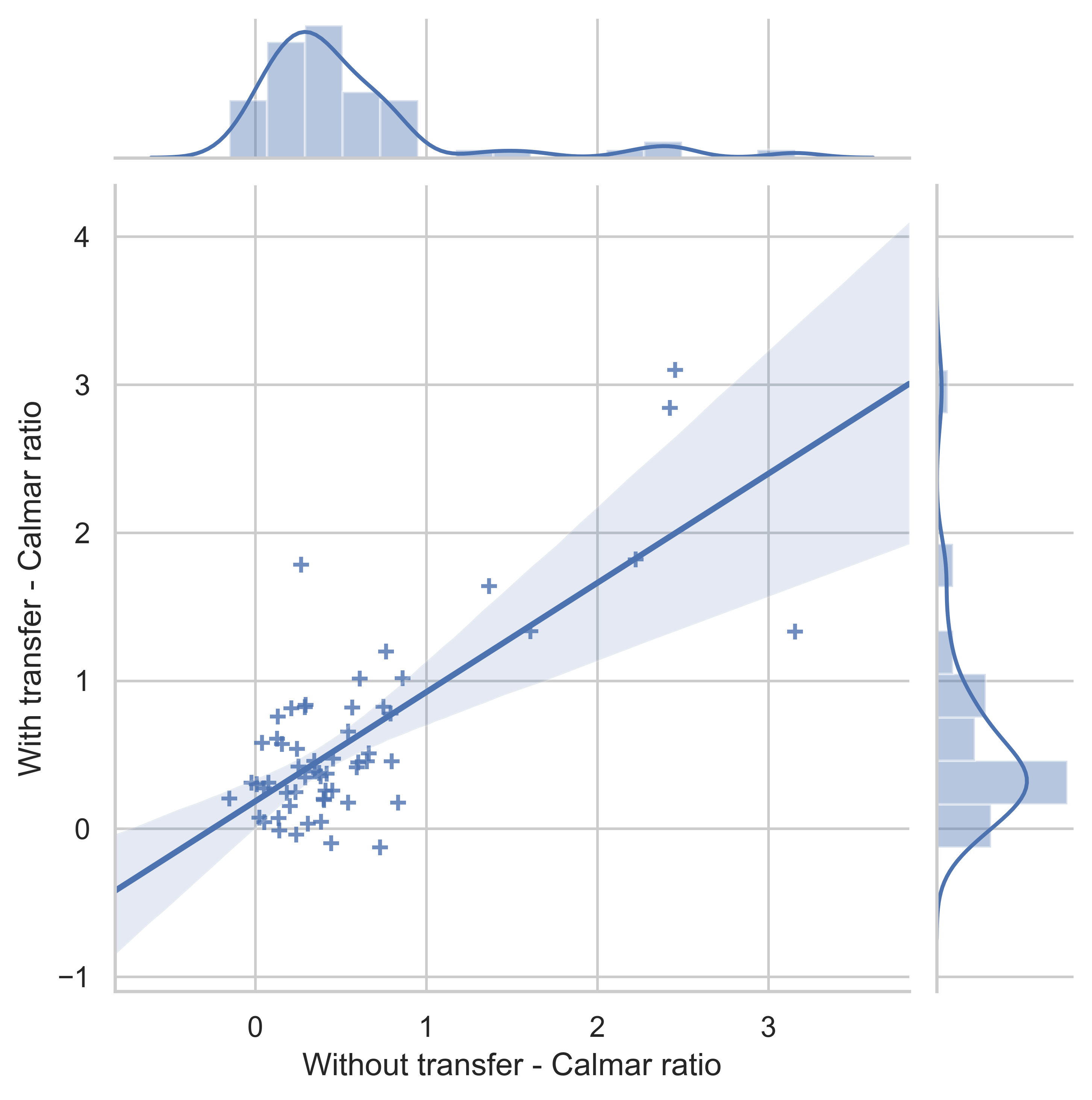}
		\caption{}
	\end{subfigure}
	\caption{Scatterplot of QuantNet and No Transfer average Sharpe (a) and Calmar (b) ratios of each market overlaid by a linear regression curve.}
	\label{fig:market-scatter-charts}
\end{figure}

In a similar fashion to the global analysis, Figures \ref{fig:market-scatter-charts}a and \ref{fig:market-scatter-charts}b display the relationship between SRs and CRs of No Transfer with QuantNet for each market, with overlaid regression curves. The SR and CR models have the following parameters: SR intercept of 0.1506 (p-value = 0.036), and SR slope of 0.7381 (p-value $<$ 0.0001); and CR intercept of 0.1851 (p-value = 0.015), and CR slope of 0.7379 (p-value $<$ 0.0001). Both cases indicate that in a market where No Transfer fared a SR or CR equal to zero, we would expect No Transfer to obtain on average 0.15 and 0.18 of SR and CR, respectively. Since both models have slope $<$ 1.0, it indicates that across markets QuantNet will tend to provide less surprisingly positive and negative SRs and CRs.

Table \ref{table:top4-markets-mean-metrics} presents a break down of the statistics in a few big regional markets, such as United States S\&P 500 components (SPX Index), United Kingdom FTSE 100 (UKX Index), Korea Composite Index (KOSPI Index) and Saudi Arabia Tadawul All Shares (SASEIDX Index). Each one of them show 2-10 times order of magnitude improvement in SRs and CRs by QuantNet, with similar benefits in Sortino ratios, Downside risks and Skewness.

\begin{table}[h!]
	\centering
	\scriptsize
	\caption{Financial metrics of QuantNet and No Transfer strategies in SPX Index, UKX Index, KOSPI Index and SASEIDX Index.}
	\label{table:top4-markets-mean-metrics}
	\begin{tabular}{c|cc|cc|cc|cc}
		\hline
		\hline
		Mean & \multicolumn{2}{c}{Americas\_SPX} & \multicolumn{2}{c}{Asia and Pacific\_KOSPI} & \multicolumn{2}{c}{Europe\_UKX} & \multicolumn{2}{c}{MEA\_SASEIDX} \\
		(SD) & No Transfer       & QuantNet      & No Transfer            & QuantNet           & No Transfer      & QuantNet     & No Transfer      & QuantNet      \\
		\hline
		Ann Ret & 0.000133          & 0.041031      & 0.073337               & 0.054152           & 0.000406         & 0.006555     & 0.000642         & 0.026092      \\
		& (0.001259)          & (0.029839)      & (0.25494)                & (0.046237)           & (0.001395)         & (0.00609)      & (0.002784)         & (0.012866)      \\
		\hline
		
		Ann Vol & 0.002436          & 0.053368      & 0.403113               & 0.07075            & 0.002717         & 0.009576     & 0.003598         & 0.01895       \\
		& (0.00064)           & (0.012918)      & (0.144194)               & (0.025008)           & (0.00111)          & (0.004583)     & (0.003219)         & (0.004739)      \\
		\hline
		
		CR       & 0.133073          & 0.756976      & 0.289853               & 0.822309           & 0.210127         & 0.813796     & 0.267246         & 1.783155      \\
		& (0.377186)          & (0.616175)      & (0.620299)               & (0.785171)           & (0.445449)         & (0.708568)     & (0.745929)         & (1.133185)      \\
		\hline
		
		DownRisk & 0.001717          & 0.035575      & 0.271677               & 0.047184           & 0.001661         & 0.005364     & 0.002236         & 0.010502      \\
		& (0.000522)          & (0.009612)      & (0.098263)               & (0.017854)           & (0.000433)         & (0.002129)     & (0.000513)         & (0.002464)      \\
		\hline
		
		Kurt     & 33.29913          & 19.51515      & 13.87472               & 22.58056           & 95.31436         & 88.42658     & 41.8963          & 33.41796      \\
		& (45.57039)          & (18.51033)      & (19.28247)               & (34.13021)           & (135.1503)         & (116.5215)     & (87.54653)         & (48.08607)      \\
		\hline
		
		MDD      & -0.00467          & -0.06681      & -0.50323               & -0.09328           & -0.00432         & -0.00988     & -0.00614         & -0.01755      \\
		& (0.002296)          & (0.028684)      & (0.190464)               & (0.051306)           & (0.001905)         & (0.005301)     & (0.003198)         & (0.007012)      \\
		\hline
		
		SR       & 0.061702          & 0.783197      & 0.347892               & 0.783428           & 0.108635         & 0.627227     & 0.12883          & 1.362307      \\
		& (0.512651)          & (0.49204)       & (0.540274)               & (0.578968)           & (0.529708)         & (0.486396)     & (0.579286)         & (0.586632)      \\
		\hline
		
		Skew     & -0.21186          & 0.28355       & 0.374017               & 0.277958           & 2.943023         & 3.573138     & 0.495534         & 2.469297      \\
		& (2.851112)          & (1.685894)      & (1.36364)                & (2.239952)           & (6.586847)         & (5.566353)     & (4.07845)          & (2.360219)      \\
		\hline
		
		SortR    & 0.144905          & 1.245084      & 0.564466               & 1.256064           & 0.301707         & 1.207481     & 0.359786         & 2.587949      \\
		& (0.760463)          & (0.847215)      & (0.839065)               & (0.979779)           & (0.898402)         & (0.979499)     & (1.206481)         & (1.344404)      \\
		\hline
		\hline
		Median & \multicolumn{2}{c}{Americas\_SPX} & \multicolumn{2}{c}{Asia and Pacific\_KOSPI} & \multicolumn{2}{c}{Europe\_UKX} & \multicolumn{2}{c}{MEA\_SASEIDX} \\
		(MAD) & No Transfer       & QuantNet      & No Transfer            & QuantNet           & No Transfer      & QuantNet     & No Transfer      & QuantNet      \\
		\hline
		Ann Ret & 0.000197          & 0.038476      & 0.059234               & 0.052772           & 0.000326         & 0.005321     & 0.000508         & 0.023614      \\
		& (0.000982)          & (0.023214)      & (0.188703)               & (0.036438)           & (0.00108)          & (0.004778)     & (0.001666)         & (0.009858)      \\
		\hline
		Ann Vol & 0.002399          & 0.053732      & 0.380229               & 0.066449           & 0.002495         & 0.008573     & 0.003128         & 0.01865       \\
		& (0.000497)          & (0.010269)      & (0.111283)               & (0.01968)            & (0.000695)         & (0.003598)     & (0.001038)         & (0.002928)      \\
		\hline
		CR       & 0.045842          & 0.626018      & 0.133249               & 0.657845           & 0.092861         & 0.654498     & 0.091411         & 1.676411      \\
		& (0.271252)          & (0.479759)      & (0.470007)               & (0.605487)           & (0.328843)         & (0.584447)     & (0.410032)         & (0.912643)      \\
		\hline
		DownRisk & 0.001648          & 0.035679      & 0.254761               & 0.044209           & 0.00157          & 0.004724     & 0.002257         & 0.009973      \\
		& (0.000395)          & (0.007424)      & (0.076679)               & (0.013787)           & (0.000333)         & (0.00173)      & (0.000405)         & (0.001975)      \\
		\hline
		Kurt     & 20.43113          & 14.44403      & 8.225536               & 13.04102           & 35.86203         & 30.40254     & 25.0625          & 28.02965      \\
		& (23.64829)          & (10.23454)      & (10.6417)                & (17.31602)           & (96.22141)         & (86.32176)     & (30.82765)         & (15.90609)      \\
		\hline
		MDD      & -0.00419          & -0.06071      & -0.49819               & -0.08132           & -0.00393         & -0.00841     & -0.00578         & -0.01573      \\
		& (0.001741)          & (0.020741)      & (0.157264)               & (0.039347)           & (0.001549)         & (0.003937)     & (0.002449)         & (0.005458)      \\
		\hline
		SR       & 0.085695          & 0.79579       & 0.366404               & 0.818361           & 0.154884         & 0.628875     & 0.184811         & 1.382302      \\
		& (0.406313)          & (0.400688)      & (0.437163)               & (0.465154)           & (0.407551)         & (0.379645)     & (0.451943)         & (0.4808)        \\
		\hline
		Skew     & 0.044563          & 0.407357      & 0.271776               & 0.148158           & 0.360259         & 1.385536     & -0.06615         & 2.559821      \\
		& (1.742953)          & (1.110312)      & (0.811512)               & (1.318841)           & (4.960393)         & (4.198536)     & (2.102935)         & (1.412129)      \\
		\hline
		SortR    & 0.112332          & 1.205009      & 0.55375                & 1.235756           & 0.203793         & 1.128096     & 0.26026          & 2.315208      \\
		& (0.595473)          & (0.690404)      & (0.675314)               & (0.779847)           & (0.701329)         & (0.793781)     & (0.764835)         & (1.124332)      \\
		\hline
		\hline
	\end{tabular}
\end{table}

\begin{figure}[h!]
	\centering
	\includegraphics[width=\textwidth]{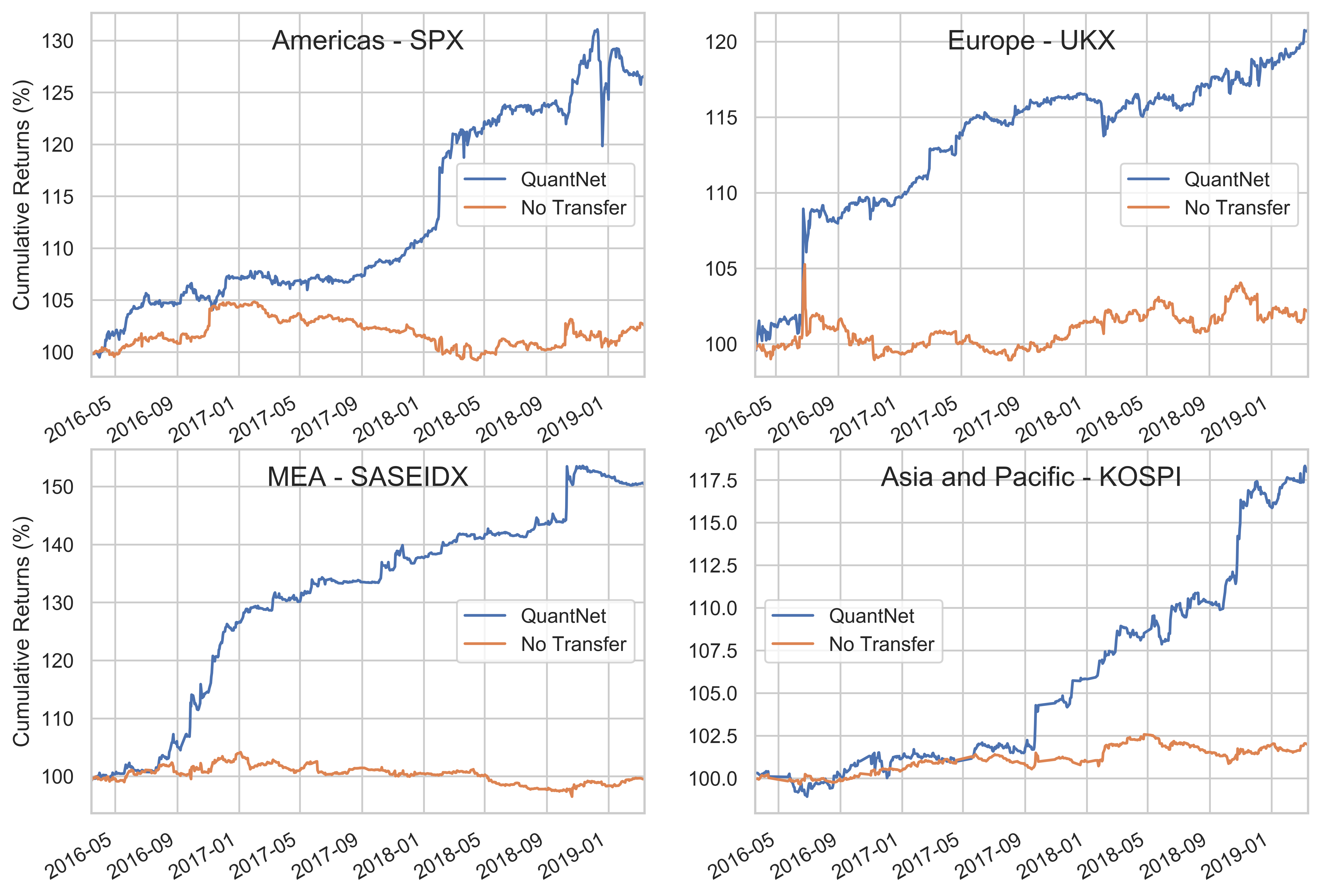}
	\caption{Average cumulative returns (\%) of SPX Index, UKX Index, KOSPI Index and SASEIDX Index contrasting QuantNet and No Transfer strategies. Before aggregation, each underlying asset was volatility-weighted to 10\%.}
	\label{fig:cumreturns-top4}
\end{figure}

\begin{figure}[h!]
	\centering
	\includegraphics[width=\textwidth]{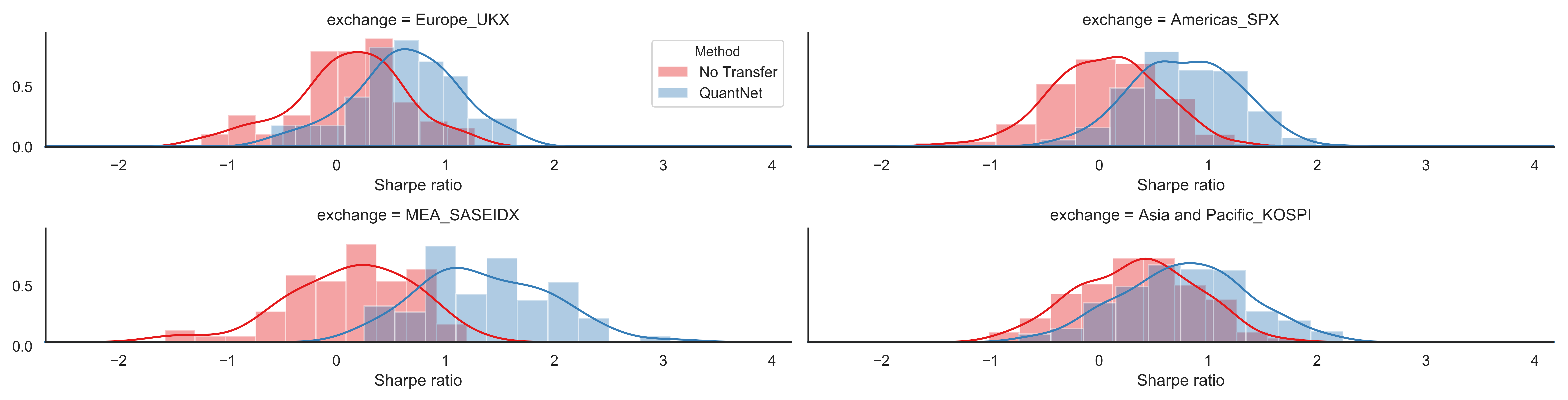}
	\caption{Histogram of Sharpe ratio of SPX Index, UKX Index, KOSPI Index and SASEIDX Index contrasting QuantNet and No Transfer strategies.}
	\label{fig:hist-top4}
\end{figure}

\begin{figure}[h!]
	\centering
	\includegraphics[width=\textwidth]{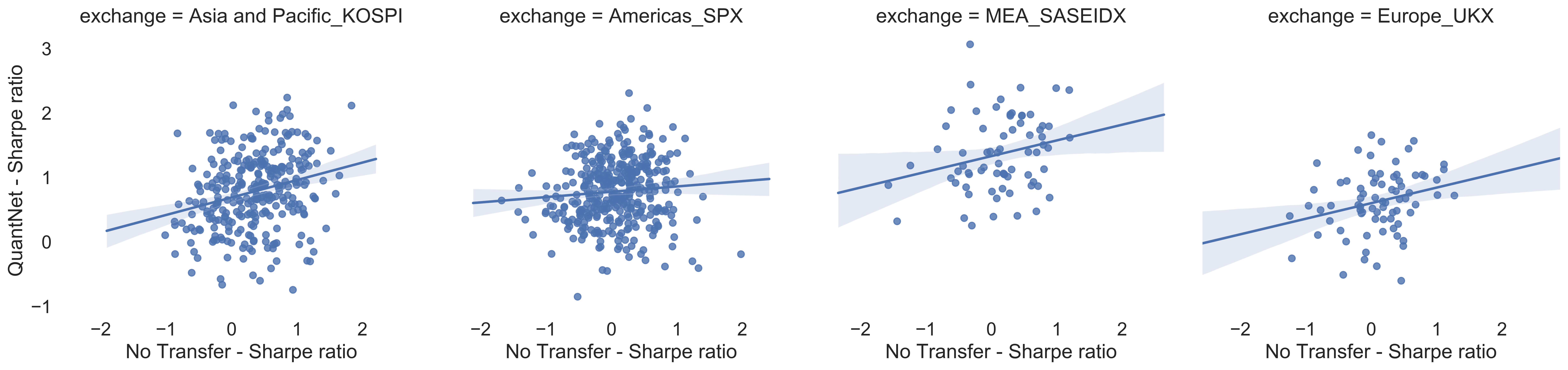}
	\caption{Scatterplot of Sharpe ratio of SPX Index, UKX Index, KOSPI Index and SASEIDX Index contrasting QuantNet and No Transfer strategies.}
	\label{fig:scatter-top4}
\end{figure}

These results by QuantNet also translate in superior cumulative returns (Figure \ref{fig:cumreturns-top4}), histograms with empirical distributions that stochastically dominate the No Transfer strategy (\ref{fig:hist-top4}), and finally positive transfer across assets (\ref{fig:scatter-top4}). In summary, markets that were otherwise not as profit-generating using only lagged information, become profitable due to the addition of a transfer layer of information across world markets. 

\clearpage

\section{Fama-French 5 Factor Model} \label{fama-french}

We fit a traditional Fama-French 5 factor model, with the addition of Momentum factor \cite{french2012kenneth,fama2015five} using these four markets daily returns as dependent variables. Table \ref{table:ff-factors} presents the models coefficients, t-stats and whether they were or not statistically significant (using a 5\% significance level). Regardless of the market, QuantNet provided significant alpha (abnormal risk-adjusted return) with very low correlation to other general market factors. 

\begin{table}[h!]
	\centering
	\caption{Models coefficients and t-stats for the different markets and factors. $*$ p-value $<$ 0.05}
	\scriptsize
	\label{table:ff-factors}
	\begin{tabular}{l|cc|cc|cc|cc}
		\hline
		\hline
		& \multicolumn{2}{c}{SPX} & \multicolumn{2}{c}{UKX} & \multicolumn{2}{c}{SASEIDX} & \multicolumn{2}{c}{KOSPI} \\
		& coefficient   & t-stat  & coefficient   & t-stat  & coefficient     & t-stat    & coefficient    & t-stat   \\
		\hline
		Alpha                                                                    & 0.0003*       & 2.554   & 0.0002*       & 2.258   & 0.0006*         & 3.856     & 0.0003*        & 4.034    \\
		Beta & -0.0004*      & -2.367  & -0.0002       & -1.558  & -0.0006*        & -2.804    & -0.0001        & -0.213   \\
		Small minus Big                                                          & 0.0003        & 1.158   & 0.0002        & 1.075   & 0.0006          & 1.856     & 0.0001         & 0.182    \\
		High minus Low                                                           & -0.0005       & -1.778  & -0.0006*      & -2.813  & -0.0002         & -0.781    & -0.0001        & -0.148   \\
		Momentum                                                                 & -0.0002       & -1.042  & 0.0001        & 0.08    & -0.0002         & -0.918    & -0.0001        & -0.538  \\
		\hline
		\hline
	\end{tabular}
\end{table}

\section{Dendrogram} \label{dendrogram}

\begin{figure}[h!]
	\centering
	\includegraphics[width=0.9\textwidth]{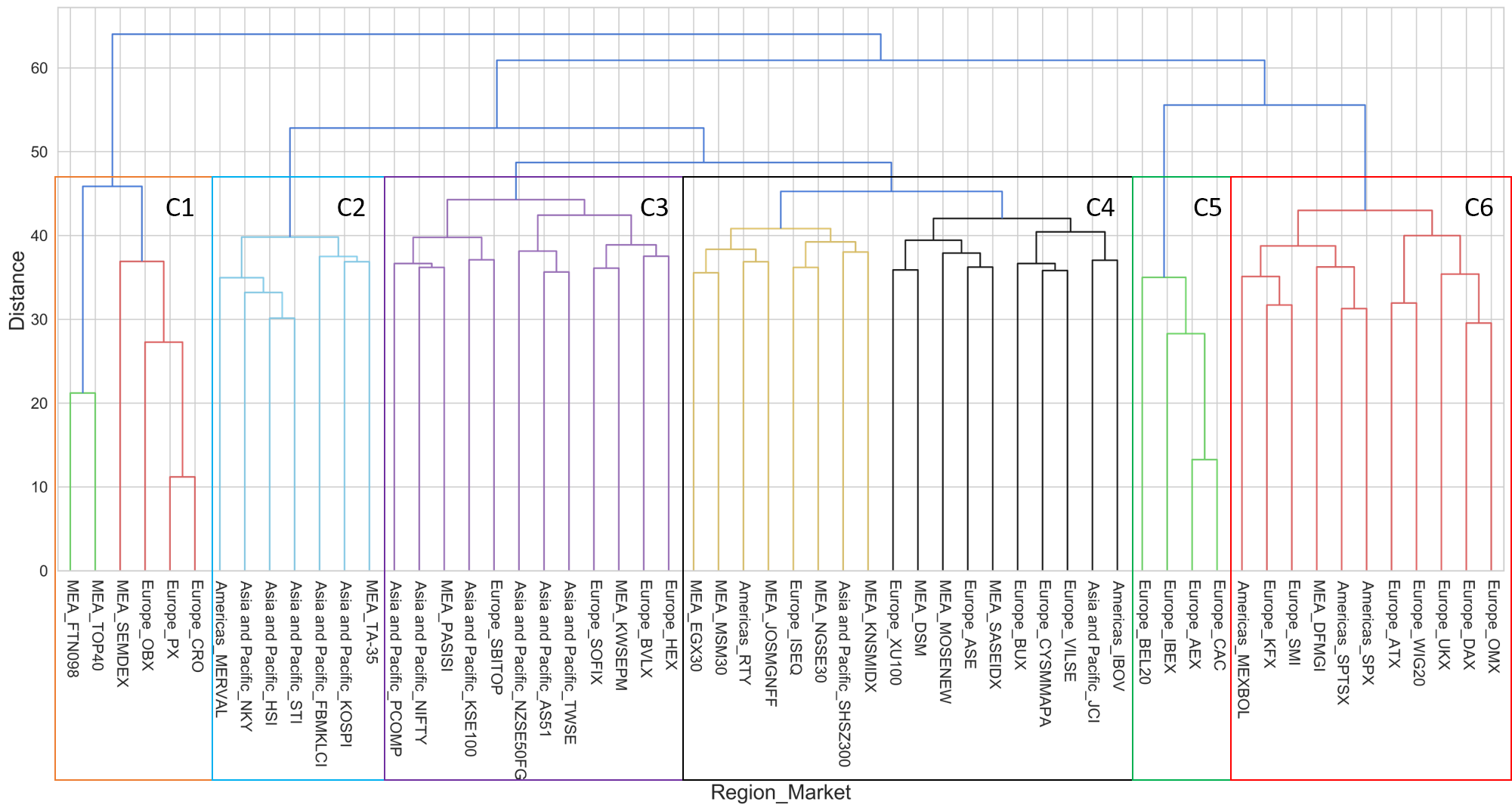}
	\caption{Dendrogram of hierarchical clustering using the scores from QuantNet encoder layer.}
	\label{fig:dendogram-encoder}
\end{figure}

An additional analysis is how each market is being mapped inside QuantNet architecture, particularly in the Encoder layer. The key question is how they are being represented in this hidden latent space, and how close each market is to the other there. Figure \ref{fig:dendogram-encoder} presents a dendrogram of hierarchical clustering done using the scores from encoder layer for all markets.

By setting an unique threshold across the hierarchical clustering we can form 6 distinct groups. Some clusters are easier to analyse, such as C5 that consist mainly of small European equity markets (Spain, Netherlands, Belgium and France) -- all neighbours; C6 comprising mainly of developed markets in Europe and Americas, such as United Kingdom, Germany, US, and their respective neighbours Austria, Poland, Switzerland, Sweden, Denmark, Canada and Mexico. Some clusters require a more refined observation, such as C2 containing most developed markets in Asia like Japan, Hong Kong, Korea and Singapore, with C3 representing Asia and Pacific emerging markets: China, India, Australia, and some respective neighbours (New Zealand, Pakistan, Philippines, Taiwan). 

\section{Ablation Study and Sensitivity Analysis} \label{ablation}

This section attempts to addresses the question: (i) could we getter better results for the No Transfer strategy; and (ii) what are the impact in QuantNet architecture by increasing its dimensionality, and performing some ablation in its architecture. Table \ref{table:ablation-dimension} presents the Sharpe ratio (SR) statistics for question (i) and (ii), by contrasting QuantNet and No Transfer strategies.

\begin{table}[h!]
	\centering
	\scriptsize
	\caption{Average Sharpe ratio per different dimensions and configurations of QuantNet and No Transfer strategies.}
	\label{table:ablation-dimension}
	\begin{tabular}{c|c|cc|cccc}
		\hline
		\hline
		Sharpe & \multirow{2}{*}{Dimension} & \multicolumn{2}{c}{No Transfer} & \multicolumn{4}{c}{QuantNet (Encoder/Decoder-Transfer Layer)}  \\
		ratio &                            & Linear            & LSTM        & Linear-Linear & Linear-LSTM & LSTM-Linear & LSTM-LSTM  \\
		\hline
		\multirow{8}{*}{Mean (SD)}    & \multirow{2}{*}{10}        & 0.324257          & 0.311424            & 0.355600      & 0.361986    & 0.370918    & 0.279758   \\
		&                           & (0.6541)          &  (0.665251)      & (0.704641)    & (0.711988)  & (0.715636)  & (0.70962) \\
		& \multirow{2}{*}{25}        & 0.324257          &  0.311424           & 0.333565      & 0.324842    & 0.325667    & 0.275786   \\
		&                            & (0.6541)          &  (0.665251)         & (0.702325)    & (0.755392)  & (0.707544)  & (0.699151) \\
		& \multirow{2}{*}{50}        & 0.324257          &  0.311424           & 0.319009      & 0.320583    & 0.234741    & 0.258272   \\
		&                            & (0.6541)          &  (0.665251)           & (0.698568)    & (0.725952)  & (0.752143)  & (0.706733) \\
		& \multirow{2}{*}{100}       & 0.324257          &  0.311424           & 0.326090      & 0.353448    & 0.228464    & 0.298445   \\
		&                            & (0.6541)          &  (0.665251)           & (0.695066)    & (0.730362)  & (0.722084)  & (0.702126) \\
		\hline
		\multirow{8}{*}{Median (MAD)} & \multirow{2}{*}{10}        & 0.306572          &  0.304244           & 0.338981      & 0.345072    & 0.354776    & 0.275548   \\
		&                            & (0.51182)         & (0.515521)            & (0.559216)    & (0.533789)  & (0.572184)  & (0.562) \\
		& \multirow{2}{*}{25}        & 0.306572          & 0.304244            & 0.314084      & 0.301769    & 0.273791    & 0.227461   \\
		&                            & (0.51182)         & (0.515521)            & (0.552154)    & (0.570677)  & (0.555298)  & (0.552615) \\
		& \multirow{2}{*}{50}        & 0.306572          & 0.304244            & 0.302167      & 0.303684    & 0.205637    & 0.219550   \\
		&                            & (0.51182)         & (0.515521)            & (0.546099)    & (0.537824)  & (0.583648)  & (0.554146) \\
		& \multirow{2}{*}{100}       & 0.306572          & 0.304244            & 0.307922      & 0.330039    & 0.188830    & 0.243308   \\
		&                            & (0.51182)         & (0.515521)            & (0.540111)    & (0.549627)  & (0.573707)  & (0.557717) \\
		\hline
		\hline
	\end{tabular}
\end{table}

In relation to No Transfer, we can perceive that there is no benefit from moving to a LSTM architecture -- in fact, we produced slightly worst outcomes in general. Maybe the lack of data per market has impacted the overall performance of this architecture. Similarly with QuantNet, a full LSTM model generated worst outcomes regardless of the dimensionality used. Linear components in QuantNet have produced better outcomes, with Linear encoders/decoders and LSTM transfer layers providing the best average results across dimensions. However, small layer sizes are linked with better SRs, and particularly for size equal to 10, the QuantNet architecture using LSTM encoders/decoders and Linear transfer layer generated the best average SRs. 

\section{Code} \label{code}
QuantNet and other strategies implementations can be found in this repository: \url{https://www.dropbox.com/sh/k7g17x5razzxxdp/AABemBvG8UI99hp14z0C8fHZa?dl=0}

\end{document}